\documentclass[letterpaper]{article} 
\usepackage{aaai25}  
\usepackage{times}  
\usepackage{helvet}  
\usepackage{courier}  
\usepackage[hyphens]{url}  
\usepackage{graphicx} 
\usepackage{natbib}  
\usepackage{caption} 
\usepackage{array}
\usepackage{longtable}
\usepackage{multirow}
\usepackage{makecell}

\usepackage{algorithm}
\usepackage{algorithmic}
\usepackage{subcaption}
\usepackage{newfloat}
\usepackage{listings}
\DeclareCaptionStyle{ruled}{labelfont=normalfont,labelsep=colon,strut=off} 
\floatstyle{ruled}
\newfloat{listing}{tb}{lst}{}
\floatname{listing}{Listing}
\usepackage{xcolor}
\newcommand{\answerYes}[1]{\textcolor{blue}{#1}} 
 
\newcommand{\answerNA}[1]{\textcolor{gray}{#1}} 
 
\title{``Harmless to You, Hurtful to Me!'': Investigating the Detection of Toxic Languages Grounded in the Perspective of Youth}

\author{
    Yaqiong Li\textsuperscript{\rm 1}, Peng Zhang\textsuperscript{\rm 1}, Lin Wang\textsuperscript{\rm 1},  Hansu Gu\textsuperscript{\rm 2}, Siyuan Qiao\textsuperscript{\rm 1}, 
    Ning Gu\textsuperscript{\rm 1},  Tun Lu\textsuperscript{\rm 1}
}
\affiliations{
    \textsuperscript{\rm 1}Fudan University, Shanghai, China
     \textsuperscript{\rm 2}Independent, Seattle, USA

    liyq22@m.fudan.edu.cn, zhangpeng\_@fudan.edu.cn, linw@fudan.edu.cn, guhansu@gmail.com, syqiao23@m.fudan.edu.cn, ninggu@fudan.edu.cn, lutun@fudan.edu.cn
}

\usepackage{bibentry}

\begin{document}
\maketitle
\begin{abstract}
Risk perception is subjective, and youth's understanding of toxic content differs from that of adults. Although previous research has conducted extensive studies on toxicity detection in social media, the investigation of youth's unique toxicity, i.e., languages perceived as nontoxic by adults but toxic as youth, is ignored. To address this gap, we aim to explore: 1) What are the features of ``youth-toxicity'' languages in social media (RQ1); 2) Can existing toxicity detection techniques accurately detect these languages (RQ2). For these questions, we took Chinese youth as the research target, constructed the first Chinese ``youth-toxicity'' dataset, and then conducted extensive analysis. Our results suggest that youth's perception of these is associated with several contextual factors, like the source of an utterance and text-related features. Incorporating these meta information into current toxicity detection methods significantly improves accuracy overall.
Finally, we propose several insights into future research on youth-centered toxicity detection.

\end{abstract}

%
\section{Introduction}
Youth, as digital natives, have been active users of social media. A 2023 Pew Research Center reported that most youth engage with TikTok (63\%), Snapchat (60\%), and Instagram (59\%), wherein 20\% describe themselves as ``always'' on TikTok. The 2023 Weibo User Report also suggested that youth users aged 16 to 21 are over 130 million, actively engaging in online discussions about entertainment, gaming, and other topics of interest. The openness of social platforms provides youth with more opportunities to disclose themselves and exchange thoughts with others, while also bringing some negative effects. A prominent one is the widespread toxic content (toxicity), defined as ``\textit{a rude, disrespectful, or unreasonable content that is likely to make someone leave a discussion}'' \cite{Perspective_11}, including hate speech \cite{Yu_13}, offensive language \cite{Zampieri_12}, harassment \cite{Mandryk_16}, unsafe sexual experiences \cite{Razi_14}, etc. 
For example, a 2022 survey revealed that nearly half of American youth (46\%) have experienced different forms of online harassment \cite{Orben_7}. Since youth is in a critical period of mental and cognitive development, exposure to toxic content is particularly dangerous as it generally triggers healthy issues like depression, eating disorders, or self-harm \cite{Shannon_5}.

For toxicity detection, researchers in Human-Computer Interaction (HCI) and Natural Language Processing (NLP) fields have made significant efforts. For example, \citet{Zampieri_12} collected various offensive language and annotated manually, based on which detection techniques driven by deep learning like BERT or LSTM, are proposed to identify toxicity from Twitter posts. \citet{Huang_13} explored the potential of Large Language Models (LLMs) like ChatGPT for hate speech detection and rationale interpretation. Additionally, \cite{Razi_14} focused on detecting unsafe sexual conversations aimed at youth, designing a machine learning-based risk detection classifier. 
These studies show significant improvements in the accuracy and efficiency of toxicity detection techniques.

However, risk perception is subjective \cite{Perhac_23}, and youth's understanding of toxic content differs from that of adults \cite{Kim_15, Park_83}. \citet{Marwick_20} stated, languages that adults view as ``bullying'' are often perceived by youth as ``drama'', especially when they involve digital conflicts or traces. Conversely, the content deemed as ``harmless'' by adults can be perceived as harmful by youth \cite{Marie_21}. Although previous research has extensively studied different types of toxic language and explored various detection techniques, a youth-centered investigation of toxicity's characteristics and detection, 
is ignored—particularly languages that are ``perceived as nontoxic by adults but toxic as youth'' (named ``youth-toxicity'' in the work). Investigating ``youth-toxicity'' languages is an urgent research topic since: 1) Youth are more vulnerable to toxic content due to their limited knowledge, experience, and cognitive capabilities; 2) These languages represent toxicity specific to youth. They are easily overlooked by general toxicity detection driven by models trained on datasets annotated by adults, having high potential risks to youth. To address this gap, we aim to explore the following questions:
\begin{itemize}
    \setlength{\topmargin}{0pt}
    \setlength{\itemsep}{0em}
    \setlength{\parskip}{0pt}
    \setlength{\parsep}{0pt}
    \item \textbf{RQ1:} What are the features of ``youth-toxicity'' languages in social media?
    \item \textbf{RQ2:} Can existing toxicity detection techniques accurately detect these languages in social media?
\end{itemize}
The exploration of the two questions faces several challenges. Current common-used toxicity datasets are not labeled from the youth's perspective and cannot adequately represent their perceptions. Thus it requires collecting a large amount of ``youth-toxicity'' utterances, which is a laborious task. Second, toxicity research in HCI and NLP fields has uncovered a broad spectrum of toxicity types, including hate speech, sexual content, etc. These types of toxicity may contain some ``youth-toxicity'' languages, further aggravating the complexity and workload of data collection. Third, there have been many kinds of detection methods such as Perspective API \cite{Perspective_11}, pre-trained language models (PLMs), and LLMs, making the investigation of RQ2 nontrivial and time-consuming. 
For the above challenges, we conducted a two-stage study. For RQ1, we developed YouthLens (a toxicity collection program), recruited 66 Chinese youth aged 13 to 21 to participate in a 15-day data collection. 5,092 ``youth-toxicity'' utterances were obtained, which are annotated with toxicity label, utterance source, toxicity type (the type comes from a systematic review of papers), and toxicity risk. To address RQ2, we considered three representative kinds of toxicity detection methods, including Perspective API, PLMs (MetaHateBERT \cite{MetaHateBERT_101}, RoBERTa\_BL \cite{SoftLabel_105}, etc.), and LLMs (GPT-4o\footnote{https://openai.com/index/hello-gpt-4o/}, Llama-3.1\footnote{https://github.com/meta-llama/llama3}, GLM-4\footnote{https://github.com/THUDM/GLM-4}, Qwen2.5\footnote{https://github.com/QwenLM/Qwen2.5}, and DeepSeek-R1\footnote{https://github.com/deepseek-ai/DeepSeek-R1}). These methods involve different released modes (open-source and closed-source models), different model sizes (PLMs and LLMs), and different languages (LLMs launched in China and abroad). 

Several findings have emerged from our detailed analysis. For RQ1, meta information such as youth attributes (age and gender) and text-related features (utterance source, text length, and LIWC semantics) are crucial factors influencing youth's perception of ``youth-toxicity'' languages. Youth are found to be more tolerant of languages from family, the significant other, or friends than those from strangers, while when ``youth-toxicity'' utterances really come from these acquaintances, especially family members, they tend to consider them as higher risks. It also suggests older and female youth are more likely to perceive utterances as ``youth-toxicity'', more sensitive to different toxic types, and more inclined to consider them as higher risk levels. In addition, several semantic features, such as specific words related to self-identity and physiological behavior, increase the likelihood that utterances are perceived as ``youth-toxicity''. For RQ2, compared to traditional methods, advanced LLMs show their potential in different ``youth-toxicity'' detection tasks, especially when informing LLMs with associated meta information. However, introducing them also brings negative effects like risk exaggeration in ``youth-toxicity'' judgment. In addition, fine-tuning can further improve LLMs' performance in detection, while the gains of few-shot learning are limited.
Overall, this study makes the following contributions. 
\begin{itemize}
    \setlength{\topmargin}{0pt}
    \setlength{\itemsep}{0em}
    \setlength{\parskip}{0pt}
    \setlength{\parsep}{0pt}
    \item To the best of our knowledge, this is the first in-depth study focusing on Chinese ``youth-toxicity'' languages.
    \item We build a corpus and investigate the features of ``youth-toxicity'' languages from several dimensions.
    \item We conduct extensive experiments to evaluate current common-used methods' performance in ``youth-toxicity'' identification.
    \item We present several insights for future research on youth-centered toxicity detection.
\end{itemize}
\section{Related Work}
\subsubsection{Online Toxic Content Encountered by Youth}
Youth are reported to be increasingly exposed to various toxic content, including cyberbullying \cite{Kim_15}, hate speech \cite{Yu_13}, and sexual content \cite{Razi_14}, etc. 
Youth are vulnerable to toxic content for two main reasons. First, growing up in a digital environment, youth are generally confident in judging online content \cite{Mcdonald_56}, while their knowledge and capability are limited in fact. Second, the complexity of online environments further increases their exposure to toxic content, such as diverse topics, broad attackers, and cross-platform migration \cite{Freed_100}.
This toxic content can cause various harms to youth \cite{Shannon_5}, such as depression, eating disorders, and self-harm. Therefore, investigating online risks from the youth's perspective and exploring the corresponding solutions have become important topics in HCI field \cite{Park_61}. 
There are some toxicity studies from the youth's perspective, such as toxicity's related features and the impact. \citet{Youth_79} conducted an analysis of risky conversations that youth encountered on Instagram. The findings indicate that risky conversations often involve sexual solicitations and mental health issues. \citet{Ali_42} further explored the features related to toxicity in risky conversations and found that metadata (e.g., conversation length and participant engagement) can better predict toxic content. 
Besides, another key finding is that the toxicity perceived by youth is not consistent with that perceived by adults \cite{Park_83}. \citet{Marwick_20} found the languages that adults consider ''bullying'' are often perceived by youth as ``drama'', especially when the languages involve digital conflicts or traces. In contrast, the content deemed as ``harmless'' by adults can be perceived as harmful by adolescents \cite{Marie_21}. These highlight the necessity of conducting toxicity research regarding different populations or groups, e.g., the characteristics of youth-perceived toxicity, which reflects the thought of human-centered computing.

\subsubsection{Toxicity Detection in Social Media}
With the growth of social platforms, toxic content has become an increasingly serious problem. Some researchers have explored automated detection methods from the perspectives of data construction and model building. Firstly, a high-quality corpus is essential for toxicity detection and researchers have released several public datasets. \citet{Elsherief_28} introduced a theoretical taxonomy of hate speech and an English benchmark corpus containing 22,056 tweets from prominent extremist
groups. \citet{MetaHateBERT_101} presented MetaHate, a meta-collection encompassing 36 hate speech English datasets. The Ciron \cite{Xiang_65} and COLD \cite{Deng_29} datasets were constructed to detect irony and offense, containing 46,180 sentences related to various topics from popular Chinese social networks. 
Previous studies have also made notable strides in detection model by leveraging techniques like machine learning, deep learning, and PLMs. 
\citet{Caselli_30} proposed a model named HateBERT for abusive language detection. It was re-trained based on BERT and achieved an improvement in performance than machine learning models.
\citet{SoftLabel_105} introduced a bi-level optimization framework (abbreviated as RoBERTa\_BL in the paper) based on RoBERTa \cite{RoBERTa_102} that combines crowdsourced annotations with the soft-labeling technique.
Recently, due to the strong understanding and reasoning capabilities, LLMs have exhibited promising performance in NLP tasks, one of which is toxicity detection \cite{Mishra_45}. \citet{Mishra_45} explored the potential of ChatGPT in detecting toxic comments on GitHub and achieved an accuracy of 60\% without fine-tuning.
However, these corpora are annotated for general toxicity without focusing on a targeted population, making the corresponding model unusable for toxicity research centered on youth. 

To bridge the gap between current toxicity research and the advocacy of human-centered computing/design, this paper aims to investigate the toxicity from the youth's perspective by taking Chinese youth as the target. It focuses on the languages ``perceived as nontoxic by adults but toxic as youth'', and employs a two-stage study. First, we constructed a Chinese dataset of ``youth-toxicity'' and analyzed its features. Then we investigated whether existing toxicity detection techniques can accurately identify them by several representative methods (Perspective API, PLMs, and LLMs).

\section{Data Collection and Analytic Methods}
This section introduces details of the ``youth-toxicity'' dataset collection and our analytic methods. 
\begin{figure}[h]
  \small
  \centering
  \setlength{\abovecaptionskip}{0.2cm}
  \setlength{\belowcaptionskip}{-0.5cm}
  \includegraphics[width=1\linewidth]{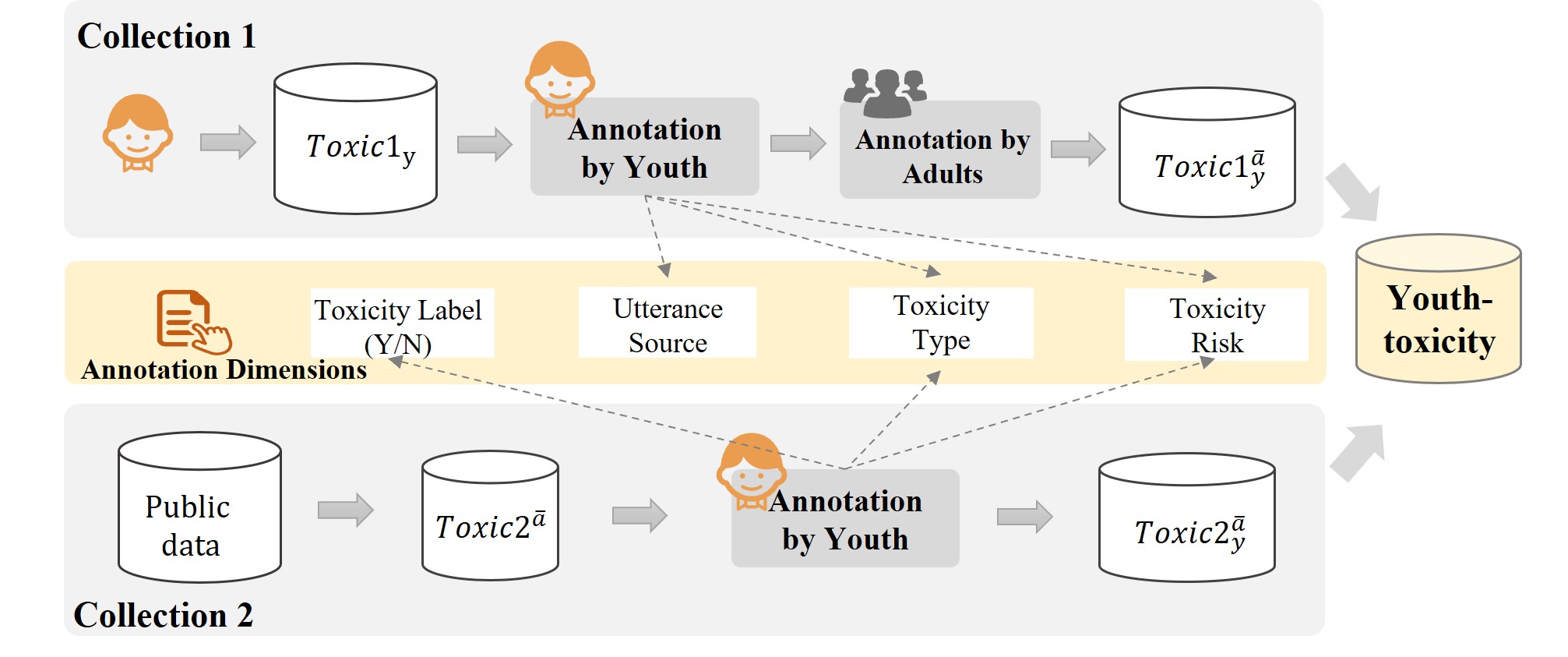}
  \caption{Data Collection Pipeline. $Toxic1_y$ represents youth-contributed toxic utterances, ${Toxic1_y^{\overline{a}}}$ denotes the youth-contributed toxic utterances but annotated as nontoxic by adults, ${Toxic2}^{\overline{a}}$ refers to nontoxic utterances within public data, and ${{Toxic2}^{\overline{a}}_y}$ denotes toxic utterances annotated by youth from ${Toxic2}^{\overline{a}}$.}
\label{fig:workflow}
\end{figure}

\subsection{Data Collection}
Existing toxicity datasets are generally annotated by adults, failing to reflect youth's perceptions. Thus, we developed a tool called YouthLens to collect ``youth-toxicity'' languages that youth encountered online. YouthLens is essentially a mini-program and involves two processes shown in Figure \ref{fig:workflow}. In Collection 1, referring to \citet{Park_83}, we encourage Chinese youth to contribute and annotate toxic content they have received from any private or public social platforms ($Toxic1_y$, the subscript $y$ indicates youth), such as WeChat and Weibo. The data obtained is then reviewed and filtered out by three authors (adults) to just remain content that they consider nontoxic (${Toxic1_y^{\overline{a}}}$, the superscript $a$ indicates adult and ``-'' indicates nontoxic). Due to the limited time and the lack of real scenarios, youth participants may struggle to recall all instances of toxic content they encountered, resulting in some bias in the collected data. To mitigate it, we introduce Collection 2, wherein each youth participant is asked to annotate 300 distinct texts from the non-toxic portions (${Toxic2}^{\overline{a}}$) of public Ciron \cite{Xiang_65} and COLD \cite{Deng_29} datasets. Since public datasets are annotated by adults, the nontoxic texts can be thought as nontoxic from adults. 
Finally, ${Toxic1_y^{\overline{a}}}$ and ${{Toxic2}^{\overline{a}}_y}$ jointly constitute the corpus for our further investigation.
In the two procedures, toxicity annotation is conducted in terms of four dimensions: toxicity label, utterance source, toxicity type, and toxicity risk.
\begin{itemize}
    \setlength{\topmargin}{0pt}
    \setlength{\itemsep}{0em}
    \setlength{\parskip}{0pt}
    \setlength{\parsep}{0pt}
    \item \textbf{Toxicity label}. Whether the current utterance contains toxicity (``Y'' indicates yes and ``N'' indicates not) based on the definition ``\textit{a rude, disrespectful, or unreasonable content that is likely to make someone leave a discussion}'' \cite{Perspective_11}. 
    \item \textbf{Utterance source}. The person the current utterance comes. Referring to \citet{Razi_24}, YouthLens offers 6 options, including ``family member'', ``significant other'', ``friend'', ``acquaintance'', ``stranger'', and ``others''. Here, ``significant other'' indicates a close partner like a boyfriend or girlfriend.
    \item \textbf{Toxicity type}. The toxicity type the current utterance belongs to. We reviewed previous toxicity studies systematically (some related to youth) and aimed to provide participants with comprehensive toxicity types. We initially selected several archives in HCI and NLP (e.g., CHI, CSCW, ICWSM, ACL) and searched for publications within the last five years (since 2020) using multiple toxicity-related keywords like ``toxic'', ``toxicity'', and ``detection'', wherein 133 papers were obtained. Three authors checked each paper's title, abstract, keywords, method, and results based on the following criteria: 1) The paper has undergone peer review and been published in a conference or journal; 2) The paper gives explicit definitions or descriptions of toxicity types studied. We got 78 papers that met these criteria, involving 138 initial toxicity types. Three authors then conducted coding of these types referring to corresponding definitions, performed a consistency check (Fleiss' Kappa = 0.901), and engaged in multiple discussions on the conflicts. Finally, seven types emerged as follows.
    \begin{itemize}
        \setlength{\topmargin}{0pt}
        \setlength{\itemsep}{0em}
        \setlength{\parskip}{0pt}
        \setlength{\parsep}{0pt}
        \item \textit{Offensive Language}: Content that is blasphemous, insulting, disgusting, morally repugnant, etc \cite{Musa_81}.
        \item \textit{Discrimination}: Content that debases individuals or groups based on race, gender, age, religion, sexual orientation, etc \cite{cignarella_77}. 
        \item \textit{Sexual Content}: Content related to the sexual topic that causes negative impacts on individuals or groups \cite{krenn_75} like sexual harassment and assault.
        \item \textit{Threat of Violence}: Content that encourages violence or attacks others, making others feel unsafe \cite{Elsherief_28}.
        \item \textit{Harassment}: Content that causes annoyance, distress, or fear, such as threats and accusations \cite{Mandryk_16}, which may be expressed repeatedly.
        \item \textit{Ideology-related Toxicity}: Toxic content that involves ideas or beliefs related to politics, society, or culture \cite{Sheth_39}. 
        \item \textit{Others}: Toxic content that is not covered in the above types but still has negative impacts \cite{Razi_24}, such as bullying and negative gossip.
    \end{itemize}
    \item \textbf{Toxicity risk}. The extent to which the utterance is likely to cause emotional or physical harm \cite{Razi_24}, is categorized as follows: \textit{Low Risk} indicates it may make the youth uncomfortable but is unlikely to cause emotional or physical harm; \textit{Medium Risk} represents it can result in harm if the youth continually encounters; \textit{High Risk} means it is deemed dangerous and cause harm. 
\end{itemize}
We recruited 66 youth aged 13 to 21, with an average age of 18.2. The age distribution was as follows: 5 participants (7.6\%) were aged 13-15, 26 participants (39.4\%) were aged 16-18, and 35 participants (53.0\%) were aged 19-21. Among the participants, 37 were male (56.1\%) and 29 were female (43.9\%). Before the annotation, we introduce to participants the definitions of toxicity, toxicity types (offensive language, discrimination, etc.), and toxicity risks (low risk, medium risk, and high risk), and then provide step-by-step instructions on how to use the YouthLens tool. Each participant is required to complete the same 50 trial annotations, and then our three authors review the results to ensure reliability. In Collection 1, youth participants' annotations cover all dimensions, and they can freely contribute any toxic content they received online without limit to the number of utterances. In Collection 2, the utterance source from public datasets is deemed as a ``stranger''. Participants are required to annotate the other three dimensions at their convenient. After completing the annotations, participants are compensated with \$15.
Finally, we collected 5,092 ``youth-toxicity'' utterances, as shown in Table \ref{tab:data statistics}, wherein 1,794 samples (${Toxic1_y^{\overline{a}}}$ in Figure \ref{fig:workflow}) were obtained via Collection 1, and 3,298 samples (${{Toxic2}^{\overline{a}}_y}$) were obtained via Collection 2. With user consent, we will release the collected data on GitHub after anonymizing private information.

\subsection{Analytic Methods}
The analysis process includes two analyses: Analysis 1 applies the logistic regression to explore the features related to ``youth-toxicity'' languages (RQ1); Analysis 2 assesses the effectiveness of detection methods in identifying these languages (RQ2), encompassing Perspective API, PLMs (MeteHateBERT, RoBERTa\_BL, etc.), and LLMs (GPT-4o, GLM-4, Qwen2.5, Llama-3.1, and DeepSeek-R1).
\setlength{\tabcolsep}{3.3pt}
\begin{table}[t]
\centering
    \centering
  \setlength{\abovecaptionskip}{0.2cm}
  \setlength{\belowcaptionskip}{-0.5cm}
    \small
    \begin{tabular}{cccc}
    \hline
    \textbf{Dimensions} & \textbf{Collection 1} & \textbf{Collection 2}  & \textbf{Total} \\ \hline
 Male & 967  & 1,782  & 2,749  \\ 
 Female & 827  & 1,516 & 2,343  \\ \hline
Aged 13-15 & 138 & 196 & 334 \\ 
  Aged 16-18 & 581 & 921 & 1,502 \\ 
  Aged 19-21 & 138 & 2,181 & 3,256 \\ 
    \hline
 Offensive Language & 1,043 & 1,548 & 2,591 \\ 
 Discrimination & 227 & 989 & 1,216 \\ 
 Sexual Content & 75 & 207 & 282 \\ 
 Threat of Violence & 42 & 66 & 108 \\
 Harassment & 113 & 69 & 182 \\ 
 Ideology-related Toxicity & 157 & 319 & 476 \\ 
 Others & 137 & 100 & 237 \\ \hline
 Family Member & 175 & 0 & 175 \\ 
 Significant Other & 46 & 0 & 46 \\   
 Friend & 268 & 0 & 268 \\   
 Acquaintance & 165 & 0 & 165 \\   
 Stranger & 963 & 3,298 & 4,261 \\   
 Others & 177 & 0 & 177 \\  \hline
 High Risk & 144 & 235 & 379  \\ 
 Medium Risk & 605 & 1,056 & 1,661 \\  
 Low Risk & 1,045 & 2,007 & 3,052 \\  \hline
    \end{tabular}
    \caption{The statistics of the ``youth-toxicity'' dataset.}
    \label{tab:data statistics}
\end{table}
\vspace{-0.2cm}
\subsubsection{Feature Analysis}
To address RQ1, we employed logistic regression to explore the features related to ``youth-toxicity'' languages. The independent variables contain youth attributes and text-related features. Given a ``youth-toxicity'' utterance, youth attributes include the age and gender of an annotator, and text-related features encompass the utterance source, text length, and LIWC semantics. LIWC semantics are extracted based on LIWC \cite{Pennebaker_62}, which is widely recognized and applied for semantic analysis in HCI and NLP research. It covers 72 categories such as ``funct'', ``pronoun'', ``social'', ``affect'', and ``cognmech'', characterizing texts from many aspects like speech, themes, emotions, and cognition. The dependent variables include toxicity label (Y/N), toxicity type, and toxicity risk. 
Based on the collection procedure, we obtained 3,588 utterances with label ``Y''. Given the balance of sample size between two labels, we randomly selected 3,588 samples with label ``N'' from Collection 2, wherein each is considered nontoxic by youth and adults. These 7,176 samples are utilized for toxicity label-oriented logistic regression analysis, and 3,588 samples with label ``Y'' are used for toxicity type or risk-oriented analysis.

Notably, features like gender and utterance source are categorical variables and need to be converted into dummy variables before the analysis. For LIWC semantics, each utterance is tokenized first, with each word represented as a 72-dimensional vector corresponding to different categories. Each dimension is assigned a value of 0 or 1, where 1 indicates that the word belongs to one category, and 0 indicates not. The utterance is then represented by the sum of its word vectors. Considering the feature co-linearity, we calculate the Spearman coefficient between any two features. If it exceeds 0.6, we remove the feature that is co-linear with more variables or less commonly used. Since toxicity type and toxicity risk are multi-category variables, the logistic analysis selects one category as the reference and transforms the others into multiple binary dependent variables. We choose ``\textit{Offensive Language}'' and ``\textit{Low Risk}'' as the reference of toxicity type and risk, respectively.

\subsubsection{Toxicity Detection}
We employed three detection methods to perform tasks for ``youth-toxicity'' languages, including toxicity label prediction, toxicity type classification, and toxicity risk classification. 
\begin{itemize}
    \setlength{\topmargin}{0pt}
    \setlength{\itemsep}{0em}
    \setlength{\parskip}{0pt}
    \setlength{\parsep}{0pt}
    \item \textbf{Perspective API}. The API, developed by Google Jigsaw, utilizes machine learning to detect toxicity and returns a score representing the toxicity level. We employed 0.7 as the threshold \cite{Perspective_11}, i.e., a text with a score above 0.7 is predicted as toxic; nontoxic otherwise. 
    \item \textbf{PLMs}. To evaluate the detection capabilities of PLMs, we selected several representative and SOTA models - MetaHateBERT \cite{MetaHateBERT_101}, RoBERTa\_BL \cite{SoftLabel_105}, HateBERT \cite{Caselli_30}, DistillBERT \cite{DistillBERT_104}, and RoBERTa \cite{RoBERTa_102} - and fine-tuned them on the collected dataset.
    \item \textbf{LLMs}. GPT-4o, GLM-4, Qwen2.5, Llama-3.1, and DeepSeek-R1 are employed. 
    They are the leading LLMs and have promising performance across multiple tasks. Given that the collected data is in Chinese, we selected three LLMs released in China (GLM-4, Qwen2.5, and DeepSeek-R1) and two advanced LLMs with multilingual language processing abilities (GPT-4o and Llama-3.1). Since Llama-3.1, GLM-4, Qwen2.5, and DeepSeek-R1 are open-source LLMs, we selected the corresponding billion-parameter versions of these LLMs for fine-tuning by referring to \citet{ModerationLLMs_107} and \citet{Distillmetahate_108}.
\end{itemize}
Among these, toxicity label prediction is achieved using the above detection methods. Since Perspective API and PLMs cannot support toxicity type classification and toxicity risk classification, these two tasks are evaluated by LLMs. Inspired by \citet{Distillmetahate_108} and \citet{ModerationLLMs_107}, we designed three kinds of prompts considering the detection role and task-specific descriptions, including the direct prompt, target-based prompt, and meta-based prompt. 
\begin{itemize}
    \setlength{\topmargin}{0pt}
    \setlength{\itemsep}{0em}
    \setlength{\parskip}{0pt}
    \setlength{\parsep}{0pt}
    \item  \textbf{Direct prompt}. Request LLMs to provide the detection result by just giving the detection role, the utterance, the task description, and the output format. 
    \item  \textbf{Target-based prompt}. In addition to the above information, provide LLMs with the targeted population to request results.
    \item \textbf{Meta-based prompt}. Tell LLMs the meta information such as the target's attributes (age and gender) and text-related features uncovered in our investigation to request results.
\end{itemize}

Through comparative analysis, we identified the best-performing prompt and further explored the impact of LLM fine-tuning and few-shot learning on its robustness. Details on prompt examples, LLM scales, and fine-tuning parameters are provided in the Appendix. All experiments were conducted on a system equipped with an Intel Xeon(R) processor and 4 NVIDIA A800 PCIe 80GB GPUs.

\section{Results}
\subsection{RQ1: Features of ``Youth-toxicity'' Language}
\subsubsection{Toxicity Label}
As shown in Table \ref{tab: LRresult}, youth's judgment on whether an utterance contains ``youth-toxicity'' is closely associated with both youth attributes and text-related features. For youth attributes, age and gender show significant positive effects, i.e., older and female youth tend to consider an utterance ``youth-toxicity''.
For text-related features, text length is a significant negative factor, indicating that shorter texts are more likely to be perceived as ``youth-toxicity''. Besides, relationships like the significant other, friend, acquaintance, and others show significant negative effects compared to strangers. This suggests that youth are less likely to view languages from familiar persons as toxic, indicating they are more tolerant of familiar people's utterances. For example, when a friend says ``\textit{You're playing really variously!}'', a youth may interpret it as a joke, while the same comment from a stranger may be seen as impolite.
For LIWC semantics, linguistic feature words like personal pronouns (``i'', ``you'', etc.), emotion-related (``anger'', ``insight'', etc.), physiological behavior (``body'', ``sexual'', and ``ingest''), social relation (``family'' and ``humans''), temporal (``time'' and ``presentM''), and special words (``multiFun'', ``negate'', and ``filler''), show varying significant associations with the toxicity label. Notably, the first-person pronoun (``i'') has a significant negative correlation with toxicity perception, while other personal pronouns exhibit a significant positive association. This suggests that the more first-person pronoun words in an utterance, the less likely it is perceived as toxic, whereas utterances with more other personal pronouns are more likely to be viewed as toxic. Pronouns like ``you'', ``she/he'', and ``they'' have strong directional characteristics, which may express aggressive attitudes to others in communication. For example, ``\textit{Come on, you don't know anything. Just listen to me!} ''
For emotion-related words, consistent with the finding of \citet{MetaHateBERT_101}, the ``anger'' type also emerges as a significant positive indicator, suggesting that utterances containing more ``anger'' words are more likely to be perceived as ``youth-toxicity''. Conversely, ``insight'', ``tentat'', ``inhib'', and ``feel'' words exhibit significant negative associations, implying that youth are more likely to view sentences with these words as nontoxic. For words related to physiological behaviors, ``body'' and ``sexual'' terms show a significant positive correlation, while ``ingest'' words have a significant negative association. This suggests that expressions involving sensitive topics related to body or sexuality are more likely to be seen as toxic, whereas sentences containing words related to food or ingestion tend to be considered nontoxic. When it comes to words related to social relationships, both ``humans'' and ``family'' terms exhibit significant positive correlations. It indicates that language referring to humans or family tends to be perceived as toxic, like a statement ``\textit{I’m your parent, and I’m criticizing you to help you become better. Why would I pick on you and not others? You should reflect on yourself.}'' 
Temporal words like ``time'' and ``presentM'' show significant negative associations, indicating utterances with these words are less likely to be seen as toxic. Some special words like ``negate'' and ``filler'' have significant positive correlations. 
Conversely, ``multiFun'' words exhibit a significant negative relationship with the toxicity label, likely due to their semantic ambiguity, making it complex to understand and reducing their likelihood to be ``youth-toxicity'' languages.
\vspace{-0.2cm}




\subsubsection{Toxicity Type}
Youth's judgment on the toxicity type of ``youth-toxicity'' (with ``\textit{Offensive Language}'' as the reference) is significantly related to youth attributes and text-related features, shown in Table \ref{tab: LRresult}. For the judgment of ``\textit{Discrimination}'', age and gender are significant negative factors, showing that younger and male youth are more likely to perceive ``youth-toxicity'' as offensive rather discrimination. The text length emerges as a significant positive factor, i.e., longer utterances tend to be classified as ``\textit{Discrimination}''. Sources such as family, significant other, friends, and acquaintances all perform as significant negative factors, meaning youth tend to perceive ``youth-toxicity'' utterances from these sources as offensive. For LIWC semantics, emotion-related (``posemo'', ``discrep'', and ``inhib'') and special words (``preps'', ``multiFun'', and ``nonfl'') are significant positive features, indicating that youth tend to interpret ``youth-toxicity'' utterances containing these words as ``\textit{Discrimination}''. Moreover, personal pronouns (``i'', ``you'', and ``ipron''), social relation (``friend''), and physiological behavior words (``achieve'' and ``death'') are also significant negative factors. Such an expression ``\textit{Who do you think you are? You can't talk to my friend like that}'' is viewed as offensive.
For the judgment of ``\textit{Sexual Content}'', gender is a significant negative factor. Female youth are more prone to interpret ``youth-toxicity'' as sexual. Besides, ``youth-toxicity'' utterances derived from friends or acquaintances are more likely to be seen as offensive, with these sources serving as significant negative factors. For LIWC semantics, personal pronouns (``they''), social relation (``friend''), physiological behavior (``body'' and ``sexual''), and special words (``preps'') are significant positive factors. The presence of these words increases the likelihood that youth perceive the content as sexual in nature, like ``\textit{Did you see what she's wearing today? It's so revealing!}''. Such words involving ``work'' and ``ipron'' are significant negative factors, indicating such utterances tend to be judged as offensive rather than sexual by youth.
For ``\textit{Threat of Violence}'', age and gender appear as significant negative factors, and longer utterances also increase the likelihood of being perceived as threatening. The source of family is a significant positive factor, meaning youth are more prone to interpret ``youth-toxicity'' from family as a threat. Additionally, the presence of personal pronouns (``i'' and ``you'') and physiological behavior words (``death'') increases the probability that youth perceive it as a threat, whereas the ``multiFun'' type like ambiguous or multi-functional words is more likely to be considered offensive. 
For ``\textit{Harassment}'', gender and text length are significant negative factors. Female youth are more likely to interpret ``youth-toxicity'' as ``\textit{Harassment}'', while longer texts are more likely to be seen as offensive. For LIWC semantics, personal pronouns (``you''), emotion-related words (``anx''), temporal words (``time''), and special words (``filler'') are positively correlated with the likelihood of ``\textit{Harassment}''. Conversely, words related to ``number'', ``excl'', and ``achieve'' are significant negative factors, suggesting ``youth-toxicity'' content containing these elements is more often seen as offensive.
For  ``\textit{Ideology-related Toxicity}'', age is a significant negative factor, meaning younger youth tend to classify it as offensive. Besides, text length and the family source are significant positive factors, indicating ``youth-toxicity'' from family members is more likely to be perceived as ``\textit{Ideology-related Toxicity}'' (stubborn ideas). For LIWC semantics, emotion-related words (``cause'', ``discrep'', and ``inhib''), temporal words (``time''), special words (``multiFun''), and physiological behavior words (``death'' and ``relig'') are positively associated with the type. Moreover, personal pronouns (``i'', ``you'', and ``youpl'') and sexual terms are significant negative features, more likely giving rise to the feeling of being offended.
For ``\textit{Others}'', sources like family, significant other, and acquaintances are significant positive factors. The ``money'' category is a significant positive factor, suggesting that youth tend to consider ``youth-toxicity'' utterances involving economic terms as ``\textit{Others}''. Those languages involving ``friend'', ``humans'', and ``space'' words are more likely to be considered offensive.

\vspace{-0.2cm}
\subsubsection{Toxicity Risk}
As shown in Table \ref{tab: LRresult}, youth's judgment of risk level of ``youth-toxicity'' languages (with ``\textit{Low Risk}'' as the reference) is significantly associated with youth attributes and text-related features. 
For the judgment of ``\textit{Medium Risk}'', age and gender are significant negative factors, meaning younger and male youth are more inclined to consider them as ``\textit{Low Risk}''. Conversely, the text length proves to be a significant positive factor. Moreover, compared to strangers, sources like family, significant other, friends, acquaintances, and others are significant positive features, suggesting that ``youth-toxicity'' from these sources tends to be more risky for youth. This could be explained by the emotional or trust-based connections between youth and these persons, so youth often expect support and understanding from them. When ``youth-toxicity'' languages come from these relationships, it may disrupt these expectations, giving rise to more negative feelings. Such utterances like ``\textit{You're too sensitive, you can't handle a bit of criticism}'' or ``\textit{I never thought you'd turn out like this}''. 
For LIWC semantics, personal pronouns (``you'') and physiological behavior words (``sexual'' and ``health'') are significant positive indicators, suggesting youth tend to perceive them as medium risk. 
For ``\textit{High Risk}'', age and gender are significant negative factors, aligning with observations in the medium risk analysis. The source of family exhibits a strong positive correlation, showing that youth are sensitive to ``youth-toxicity'' words from family members and tend to perceive them as high risks. Moreover, ``presentM'' and ``sexual'' terms are significant positive features. This indicates that youth incline to consider discussions involving sexual topics high risk.


\textbf{We obtain the following conclusions.} For youth attributes, older and female youth are more likely to perceive utterances as ``youth-toxicity'', more sensitive to its types, and perceive it as higher risks. Conversely, younger and male youth tend to view ``youth-toxicity'' as ``\textit{Offensive Language}'' with low risk. For utterance source, youth exhibit higher tolerance for utterances from non-strangers like family, the significant other, and friends, i.e., such languages are less likely to be perceived as toxic. However, when ``youth-toxicity'' utterances from these sources, youth tend to consider them as higher risks. Such languages from family are more likely to be treated as a threat with ``\textit{High Risk}'', and those from others are also likely to be considered ``\textit{Medium Risk}'' offensive languages. Besides, shorter texts are more likely to be viewed as ``youth-toxicity'', and youth are more sensitive to specific words like personal pronouns (``you'' and ``she/he''), social relation (``family'', ``friend'', and ``humans''), physiological behavior (``sexual'' and ``body''), and special terms (``negate'' and ``filler''). Discussions involving sexual topics are more prone to be viewed as high risk.

\subsection{RQ2: Performance of Toxicity Detection Methods}

\subsubsection{Toxicity Label Prediction}
For Perspective API, the detection accuracy in the toxicity label prediction is 0.440, with an F1-score of 0.276 (see Table \ref{tab:detection_baselines}). Conversely, fine-tuned PLMs achieve higher accuracy and F1-scores compared to the Perspective API. Specifically, accuracy ranges from 0.566 (MetahateBERT) to 0.573 (DistillBERT), 0.592 (HateBERT), 0.605 (RoBERTa), and 0.613 (RoBERTa\_RL), while F1-scores are 0.721 (MetahateBERT), 0.627 (DistillBERT), 0.645 (HateBERT), 0.659 (RoBERTa), and 0.679 (RoBERTa\_RL), respectively.
\begin{table}[htbp]
\centering
  \setlength{\abovecaptionskip}{0.2cm}
  \setlength{\belowcaptionskip}{-0.2cm}
\small
\begin{tabular}{ccccc}
\hline
\textbf{Method}   & \textbf{Precision}& \textbf{Recall}& \textbf{F1-score}& \textbf{Accuracy}\\ \hline
Perspective API  & 0.438      & 0.505           & 0.276     & 0.440   \\ 
RoBERTa  & 0.644      & 0.674           & 0.659     & 0.605   \\ 
HateBERT  & 0.636      & 0.653           & 0.645     & 0.592   \\ 
DistillBERT  & 0.620     & 0.634           & 0.627     & 0.573   \\ 
MetahateBERT & 0.567      & 0.990           & 0.721     & 0.566   \\ 
RoBERTa\_BL  & 0.640      & 0.721           & 0.679     & \textbf{0.613}  \\ \hline
\end{tabular}
\caption{The performance of baseline models.} 
\label{tab:detection_baselines}
\end{table}
\vspace{-0.2cm}

\begin{figure*}[!h]
  \small
  \centering
  \begin{minipage}{0.49\linewidth}
    \centering
    \includegraphics[width=\linewidth]{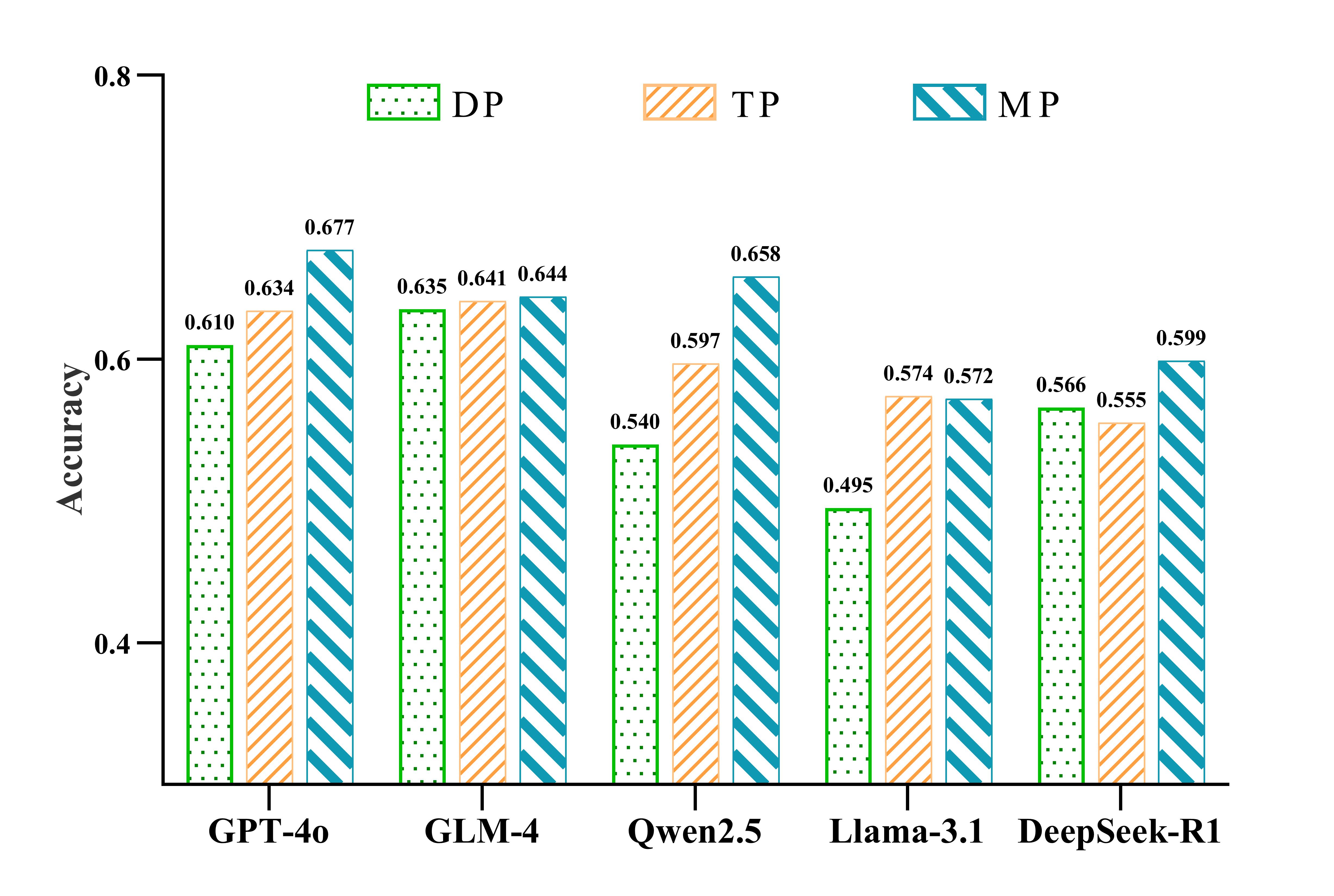}
    \subcaption{Accuracy of LLMs on toxicity label prediction}
  \end{minipage}
  \hfill
  \begin{minipage}{0.49\linewidth}
    \centering
    \includegraphics[width=\linewidth]{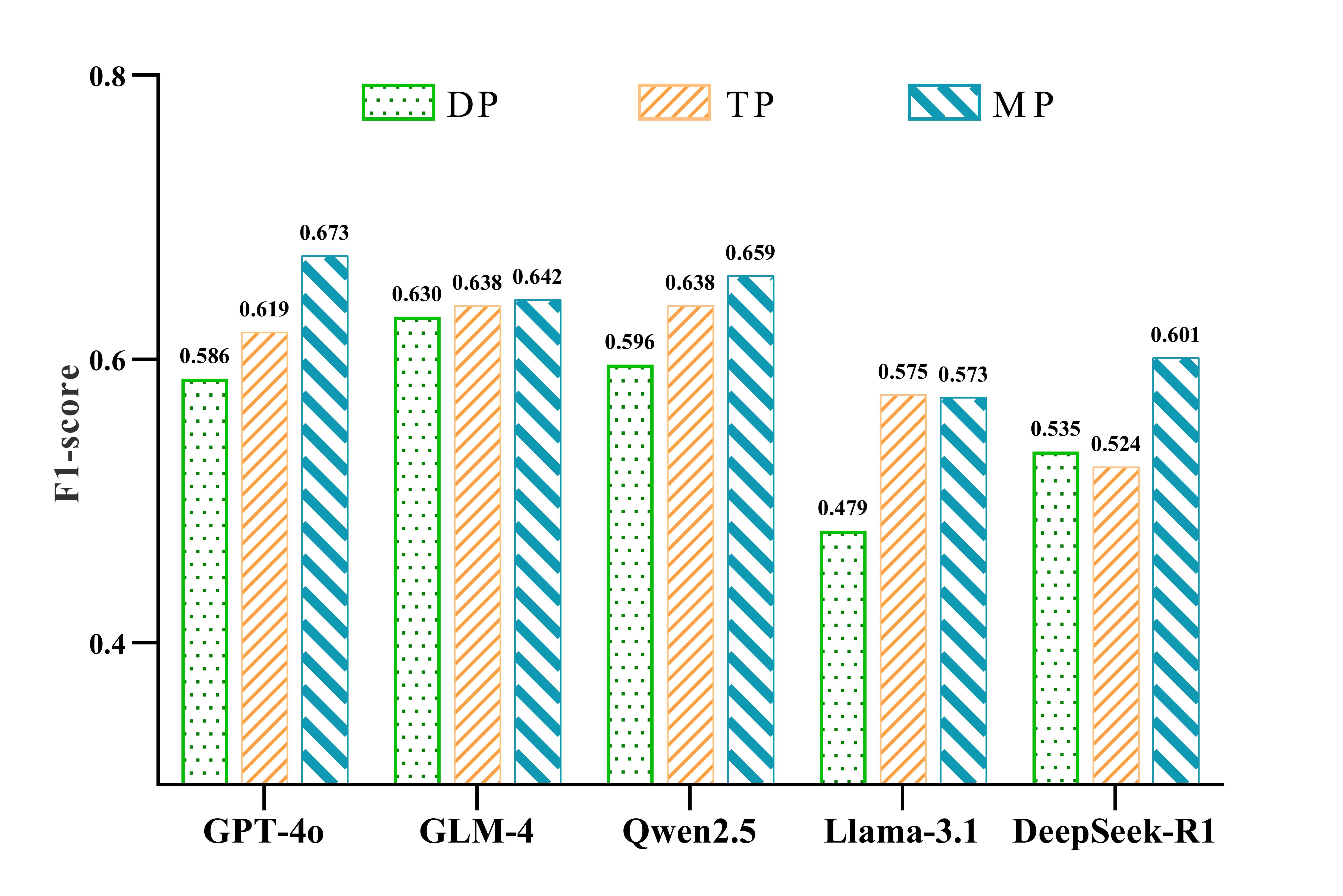}
    \subcaption{F1-scores of LLMs on toxicity label prediction}
  \end{minipage}
  \caption{LLM performance on toxicity label prediction using different prompts. DP refers to Direct Prompt, TP refers to Target-based Prompt, and MP refers to Meta-based Prompt.}
    \label{fig:toxicity_label}
\end{figure*}

For LLMs, as shown in Figure \ref{fig:toxicity_label}(a), the detection accuracy of most models across different prompts exceeds 0.5. With the direct prompt, accuracy ranges from 0.495 (Llama-3.1) to  0.540 (Qwen2.5), 0.566 (DeepSeek-R1), 0.610 (GPT-4o), and 0.635 (GLM-4), indicating a slight performance difference among LLMs. Similarly, each LLM performs dynamically with different prompts. The detection accuracy of GPT-4o, GLM-4, Qwen2.5, and Llama-3.1 with the target-based prompt improves by 2.4\%, 0.6\%, 5.7\%, and 7.9\%, respectively, compared to the corresponding accuracy of the direct prompt. These results suggest that by informing LLMs of the target audience, their capability to detect ``youth-toxicity'' can be enhanced to some extent. With the meta-based prompt, the accuracy of GPT-4o, GLM-4, Qwen2.5, Llama-3.1, and DeepSeek-R1 improves more prominently, with the improvement of 6.7\%, 0.9\%, 11.8\%, 7.7\%, and 3.3\%, respectively, compared with the corresponding accuracy of direct prompt. This indicates that informing LLMs of the meta information related to ``youth-toxicity'' is more effective in promoting detection accuracy. The changes of F1-scores shown in Figure \ref{fig:toxicity_label}(b) also overall align with the evolution of accuracy, further validating the above conclusion.
We also conducted an error analysis to see what kinds of utterances cannot be correctly identified in the ``youth-toxicity'' label prediction task. 
Referring to \citet{Ali_42}, we chose the best-performing LLM (GPT-4o) as the representative, focusing on two kinds of errors: False Negative (FN) and False Positive (FP). For FN samples, which means real ``youth-toxicity'' utterances that are mistakenly classified as nontoxic, were found to be mostly shorter texts with implicit expressions, e.g., expressing sarcasm implicitly. Like ``\textit{Actually, everyone knows whether you’ve worked hard or not}'' subtly criticizes the youth's insufficient effort without using any explicit negative terms. 
For FP samples, i.e., the nontoxic utterances that are incorrectly classified as ``youth-toxicity'', most involve expressions of emotions, jokes, or opinions. An emotional catharsis like ``\textit{Great, been crawling around in the shadows for a long time}'', is mistakenly identified. 
Besides, opinions like ``\textit{it is because of the so-called feminists that many women are disgusted with women's rights}'' are related to sensitive topics like women, potentially leading the LLM to mistakenly judge them as discrimination.

In addition, the results show that the meta-based prompt achieves the best performance overall. Thus, we further analyzed the impact of LLM fine-tuning and few-shot learning on this prompt. As shown in Table \ref{tab:detection_FT}, the accuracy of LLMs without fine-tuning is 0.644 (GLM-4), 0.658 (Qwen2.5), 0.572 (Llama-3.1), and 0.599 (DeepSeek-R1), respectively. After fine-tuning, the accuracy of each LLM becomes 0.732 (GLM-4), 0.660 (Qwen2.5), 0.632 (Llama-3.1), and 0.563 (DeepSeek-R1), respectively. This indicates that fine-tuning improves toxicity label detection for most LLMs, except for DeepSeek-R1. It may be attributed to its ``over-thinking'' issue \cite{Overthinking_109} caused by excessive input context. 
According to Table \ref{tab:detection_fewshot}, accuracy of LLMs using few-shot examples is 0.700 (GPT-4o), 0.651 (GLM-4), 0.622 (Qwen2.5), 0.482 (Llama-3.1), and 0.614 (DeepSeek-R1), respectively. It suggests that few-shot technique can benefit most LLMs like GPT-4o, GLM-4, and DeepSeek-R1 in toxicity detection.

\begin{figure*}[!h]
  \small
  \centering
  \begin{minipage}{0.49\linewidth}
    \centering
    \includegraphics[width=\linewidth]{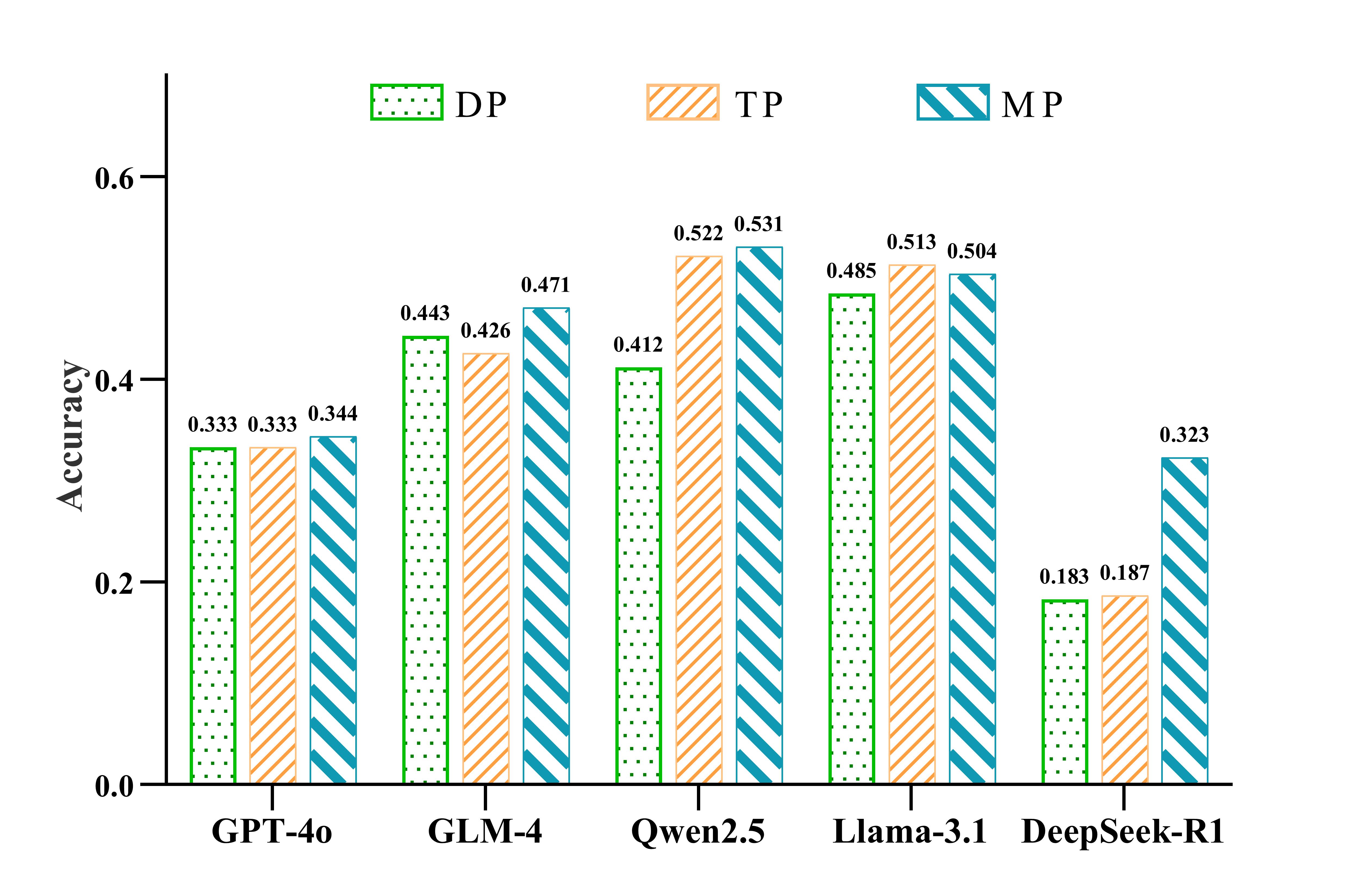}
    \subcaption{Accuracy of LLMs on toxicity type classification}
  \end{minipage}
  \hfill
  \begin{minipage}{0.49\linewidth}
    \centering
    \includegraphics[width=\linewidth]{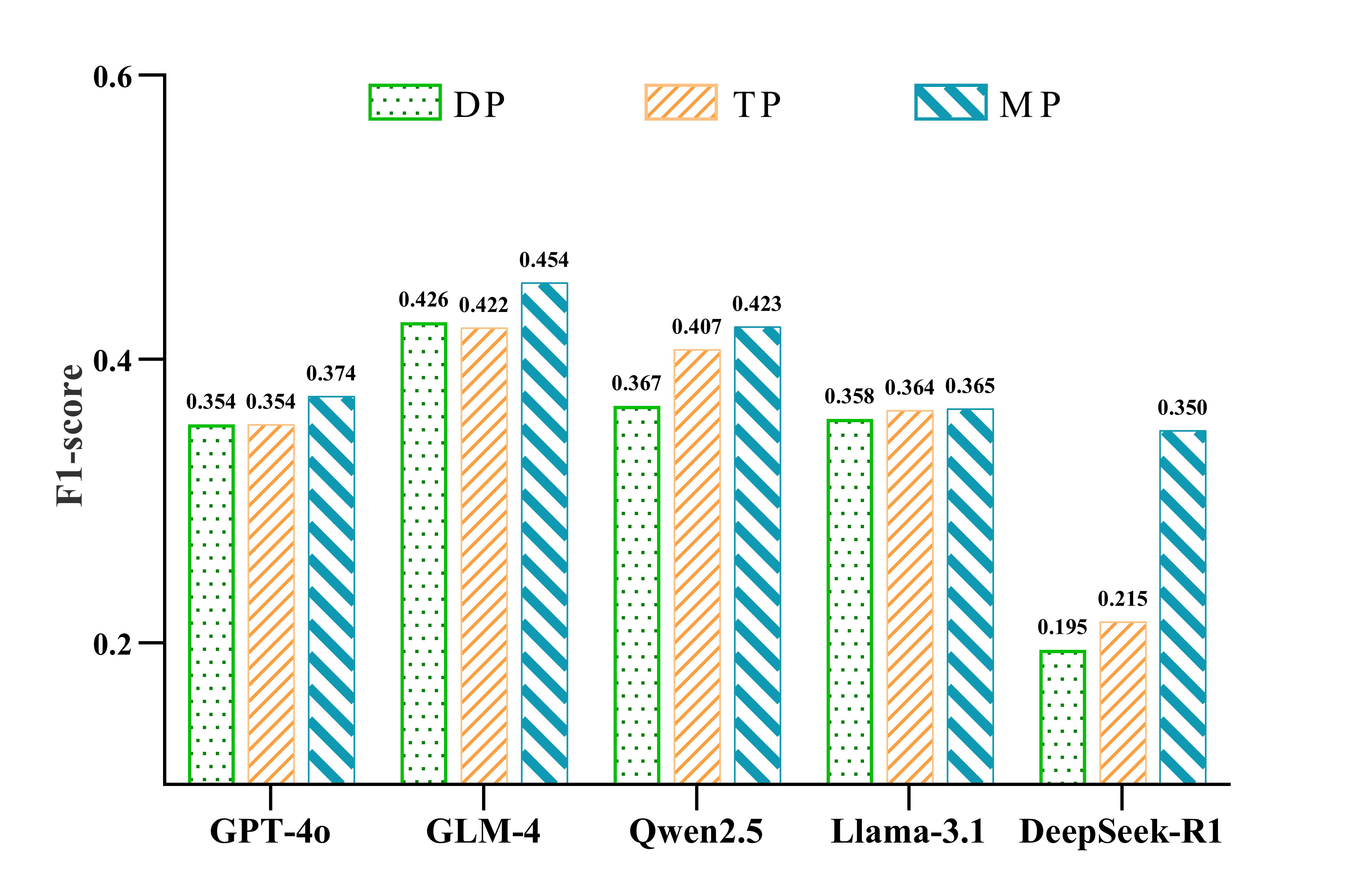}
    \subcaption{F1-scores of LLMs on toxicity type classification}
  \end{minipage}
  \caption{LLM performance on toxicity type classification using different prompts (using the same representations as Figure \ref{fig:toxicity_label}).}
    \label{fig:toxicity_type}
\end{figure*}

\begin{figure*}[!h]
  \small
  \centering
  \begin{minipage}{0.49\linewidth}
    \centering
    \includegraphics[width=\linewidth]{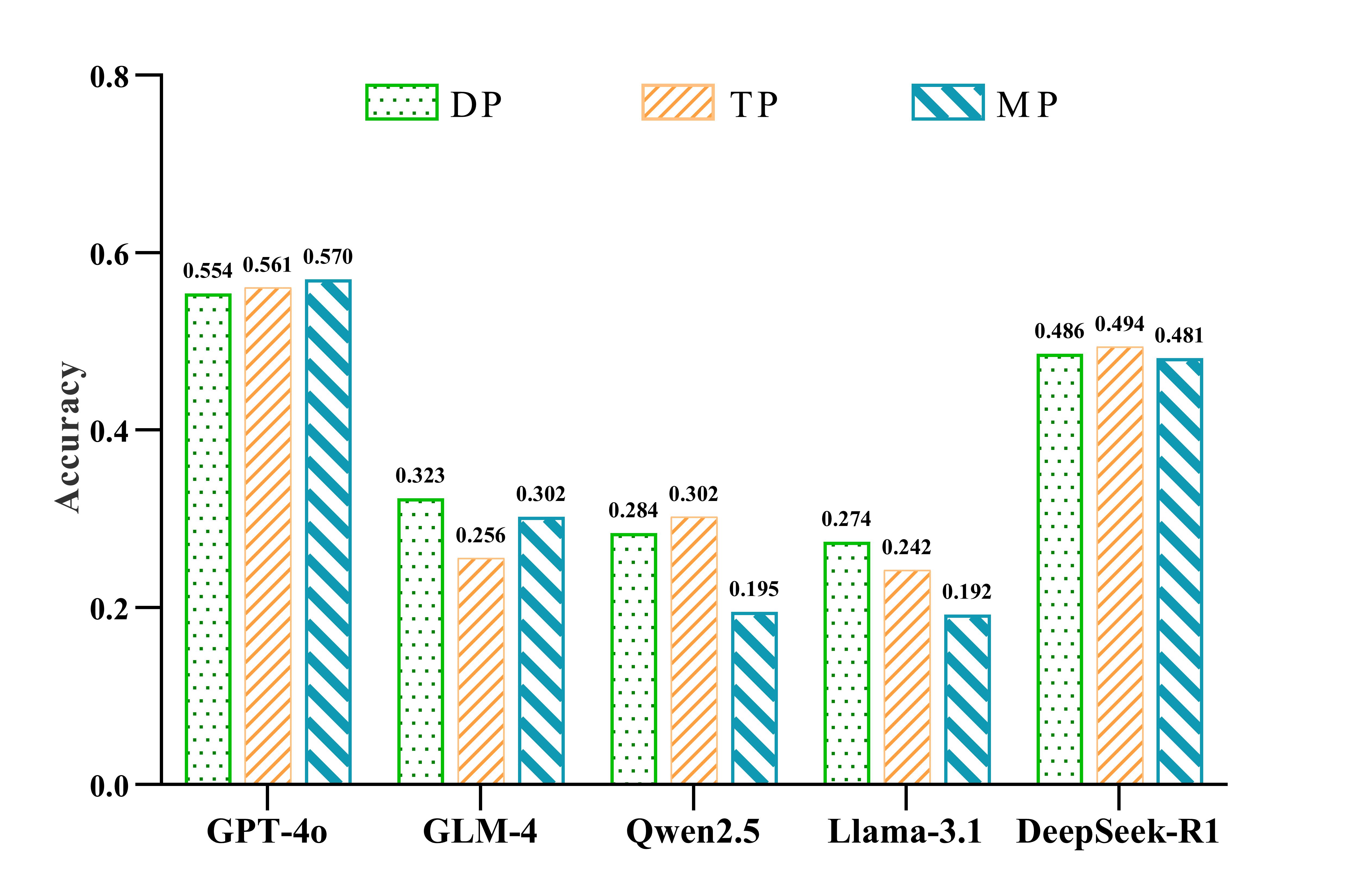}
    \subcaption{Accuracy of LLMs on toxicity risk classification}
  \end{minipage}
  \hfill
  \begin{minipage}{0.49\linewidth}
    \centering
    \includegraphics[width=\linewidth]{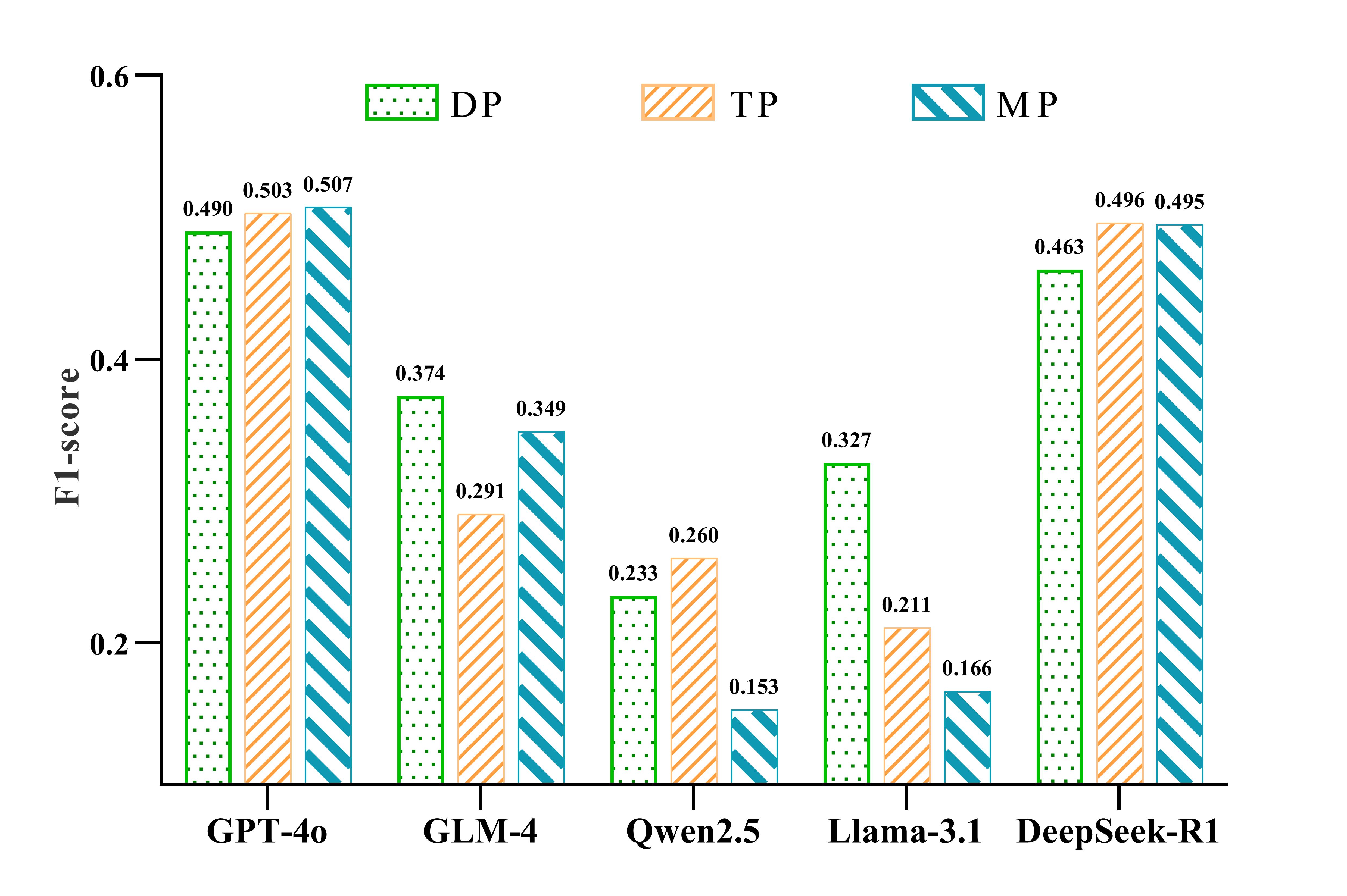}
    \subcaption{F1-scores of LLMs on toxicity risk classification}
  \end{minipage}
  \caption{LLM performance on toxicity risk classification using different prompts (using the same representations as Figure \ref{fig:toxicity_label}).}
    \label{fig:toxicity_risk}
\end{figure*}
\vspace{-0.2cm}
\subsubsection{Toxicity Type Classification}
As illustrated in Figure \ref{fig:toxicity_type}, LLMs' performance in identifying the toxicity type of ``youth-toxicity'' varies significantly. With the direct prompt, detection accuracy of LLMs ranges from 0.183 (DeepSeek-R1) to 0.333 (GPT-4o), 0.412 (Qwen2.5), 0.443 (GLM-4), and 0.485 (Llama-3.1). This trend is also observed in the use of target-based or meta-based prompt. Additionally, as detailed in Table \ref{tab:type_detection}, the performance also varies in terms of different types of ``youth-toxicity''. For example, using the direct prompt, GLM-4 and Qwen2.5 achieve the highest F1-score in detecting ``\textit{Offensive Language}'', with 0.596 and 0.592, respectively, while the score is lower in identifying ``\textit{Discrimination}'' (0.398 and 0.258, respectively). 
When varying the prompt, compared to the direct prompt, the accuracy of DeepSeek-R1, Llama-3.1, and Qwen2.5 with the target-based prompt improves by 0.4\%, 2.8\%, and 11.0\%, respectively. Only Qwen2.5's improvement is greater than 10\%, and the gains of the other LLMs are not significant. This suggests that informing LLMs of the target audience (youth) cannot significantly improve the performance of toxicity type identification. When using the meta-based prompt, detection accuracy of GPT-4o, GLM-4, Qwen2.5, Llama-3.1, and DeepSeek-R1 increases to 0.344, 0.471, 0.531, 0.504, and 0.323, respectively. Compared to the direct prompt, the meta-based prompt results in a more prominent performance improvement, indicating that providing the youth's attributes and features related to the toxicity type is more helpful. This trend is also reflected in the F1-score shown in Figure \ref{fig:toxicity_type}(b). 
Furthermore, different degrees of improvement are observed in terms of different types in Table \ref{tab:type_detection} when adopting the meta-based prompt. The improvements are particularly notable for ``\textit{Offensive Language}'' and ``\textit{Discrimination}'' types. For example, GLM-4 shows improvements across most types except for ``\textit{Threat of Violence}'', wherein the most significant gain is in ``\textit{Sexual Content}'', with the F1-score increasing from 0.241 to 0.323. GPT-4o and DeepSeek-R1 also show improvements in detecting multiple toxicity types. The F1-score of GPT-4o for ``\textit{Ideology-related Toxicity}'' increases from 0.163 to 0.297. Moreover, the F1-score of DeepSeek-R1 for ``\textit{Offensive Language}'' increases from 0.194 to 0.460, and that for ``\textit{Sexual Content}'' increases from 0 to 0.233.
\begin{table}[htbp]
\centering
  \setlength{\abovecaptionskip}{0.2cm}
  \setlength{\belowcaptionskip}{-0.2cm}
\small
\begin{tabular}{ccccc}
\hline
\multicolumn{1}{c}{\multirow{2}{*}{\textbf{LLMs}}} & \multicolumn{2}{c}{\textbf{Without Fine-tuning}} & \multicolumn{2}{c}{\textbf{With Fine-tuning}} \\
   & \textbf{Accuracy}& \textbf{F1-score}& \textbf{Accuracy}& \textbf{F1-score}\\ \hline
\multicolumn{5}{c}{Toxicity Label Prediction} \\ \hline
GLM-4 & 0.644 & 0.641 & \textbf{0.732} & \textbf{0.731}   \\
Qwen2.5 & 0.658 & 0.659 & \textbf{0.660} & 0.658   \\
Llama-3.1 & 0.572 & 0.573 & \textbf{0.632} & \textbf{0.610}   \\
DeepSeek-R1 & 0.599 & 0.601 & 0.563 & 0.545   \\ \hline
\multicolumn{5}{c}{Toxicity Type Classification} \\ \hline
GLM-4 & 0.471 & 0.454 & \textbf{0.545} & 0.428   \\
Qwen2.5 & 0.531 & 0.423 & \textbf{0.532} & \textbf{0.426}   \\
Llama-3.1 & 0.504 & 0.365 & \textbf{0.520} & 0.355   \\
DeepSeek-R1 & 0.323 & 0.350 & 0.256 & 0.286   \\ \hline
\multicolumn{5}{c}{Toxicity Risk Classification} \\ \hline
GLM-4 & 0.302 & 0.349 & \textbf{0.591} & \textbf{0.538}   \\
Qwen2.5 & 0.135 & 0.153 & \textbf{0.591} & \textbf{0.499}   \\
Llama3.1 & 0.192 & 0.166 & \textbf{0.520} & \textbf{0.500}   \\
DeepSeek-R1 & 0.481 & 0.495 & \textbf{0.489} & \textbf{0.498}   \\  \hline
\end{tabular}
\caption{The performance of LLMs with and without fine-tuning using meta-based prompt.} 
\label{tab:detection_FT}
\end{table}

\vspace{-0.2cm}
As shown in Table \ref{tab:detection_FT}, the accuracy of LLMs without fine-tuning is 0.471 (GLM-4), 0.531 (Qwen2.5), 0.504 (Llama-3.1), and 0.323 (DeepSeek-R1), respectively. After fine-tuning, the accuracy is 0.545 (GLM-4), 0.532 (Qwen2.5), 0.520 (Llama-3.1), and 0.256 (DeepSeek-R1), respectively. 
This observation is consistent with the finding in the toxicity label prediction task: fine-tuning improves the detection performance of most LLMs except for DeepSeek-R1. Besides, after introducing few-shot examples, the accuracy drops to 0.341 (GPT-4o), 0.440 (GLM-4), 0.481 (Qwen2.5), 0.476 (Llama-3.1), and 0.220 (DeepSeek-R1), respectively. It indicates that the few-shot technique does not lead to performance gains in the toxicity type classification.

\subsubsection{Toxicity Risk Classification}
The results for toxicity risk classification are shown in Figure \ref{fig:toxicity_risk}. When using the direct prompt, detection accuracy ranges from 0.274 (Llama-3.1) to 0.284 (Qwen2.5), 0.323 (GLM-4), 0.486 (DeepSeek-R1), and 0.554 (GPT-4o). This trend is also reflected in the results of the meta-based prompt. As detailed in Table \ref{tab:risk_detection}, LLMs perform better in identifying the low risk. For three prompts, GPT-4o and DeepSeek-R1 achieve F1-scores higher than 0.6 for the low risk level, outperforming the corresponding scores of other risks. GLM-4 and Qwen2.5 also get higher F1-scores in the low risk level than others.
For different prompts, the accuracy of GPT-4o using the target-based and meta-based prompt is 0.561 and 0.570, respectively, showing slight variation compared to the accuracy (0.554) of the direct prompt. Varying prompts from direct to target-based and meta-based prompt, GLM-4's accuracy changes from 0.323 to 0.256 and 0.302, respectively, demonstrating a decreasing trend. Similar changes are observed in the results of Qwen2.5 and Llama-3.1, indicating that the latter two prompts cannot help LLMs improve the performance in identifying toxicity risk. We further split results based on risk levels to understand the decreasing trend in detail. The performance of low risk identification declines significantly after adopting the latter two prompts. For example, GLM-4 achieves an F1-score of 0.437 for low risk identification using the direct prompt, while the score drops to 0.421 and 0.412, respectively, when adopting target-based and meta-based prompt. A major reason is that when informing LLMs with the target or meta information, LLMs become more strict with the toxicity risk and tend to exaggerate the low risk to higher levels. When applying the meta-based prompt to GLM-4, the number of real low risk samples misclassified as medium or high risk increases from 1,170 to 1,275, and medium risk samples misclassified as high risk rise from 175 to 275. 
\begin{table}[htbp]
\centering
  \setlength{\abovecaptionskip}{0.2cm}
  \setlength{\belowcaptionskip}{-0.2cm}
\small
\begin{tabular}{ccccc}
\hline
\multicolumn{1}{c}{\multirow{2}{*}{\textbf{LLMs}}} & \multicolumn{2}{c}{\textbf{Without Few-shot}} & \multicolumn{2}{c}{\textbf{With Few-shot}} \\
   & \textbf{Accuracy}& \textbf{F1-score}& \textbf{Accuracy}& \textbf{F1-score}\\ \hline
\multicolumn{5}{c}{Toxicity Label Prediction} \\ \hline
GPT-4o & 0.677 & 0.673 & \textbf{0.700} & \textbf{0.699}   \\
GLM-4 & 0.644 & 0.641 & \textbf{0.651} & \textbf{0.651}   \\
Qwen2.5 & 0.658 & 0.659 & 0.622 & 0.610   \\
Llama-3.1 & 0.572 & 0.573 & 0.482 & 0.375   \\
DeepSeek-R1 & 0.599 & 0.601 & \textbf{0.614} & \textbf{0.615}   \\ \hline
\multicolumn{5}{c}{Toxicity Type Classification} \\ \hline
GPT-4o & 0.344 & 0.374 & 0.341 & 0.364   \\
GLM-4 & 0.471 & 0.454 & 0.440 & 0.370   \\
Qwen2.5 & 0.531 & 0.423 & 0.481 & 0.403   \\
Llama-3.1 & 0.504 & 0.365 & 0.476 & 0.361   \\
DeepSeek-R1 & 0.323 & 0.350 & 0.220 & 0.252   \\  \hline
\multicolumn{5}{c}{Toxicity Risk Classification} \\ \hline
GPT-4o & 0.570 & 0.507 & 0.566 & \textbf{0.537}   \\
GLM-4 & 0.302 & 0.349 & \textbf{0.364} & \textbf{0.411}   \\
Qwen2.5 & 0.135 & 0.153 & \textbf{0.284} & \textbf{0.326}   \\
Llama-3.1 & 0.192 & 0.166 & 0.124 & 0.143   \\
DeepSeek-R1 & 0.481 & 0.495 & 0.411 & 0.449   \\ \hline
\end{tabular}
\caption{The performance of LLMs with and without few-shot examples using meta-based prompt.} 
\label{tab:detection_fewshot}
\end{table}
\vspace{-0.2cm}

Moreover, the accuracy of LLMs without fine-tuning is 0.302 (GLM-4), 0.135 (Qwen2.5), 0.192 (Llama-3.1), and 0.481 (DeepSeek-R1), respectively. After fine-tuning, the accuracy is 0.591 (GLM-4), 0.591 (Qwen2.5), 0.520 (Llama-3.1), and 0.489 (DeepSeek-R1), respectively. It indicates that fine-tuning significantly improves the performance of LLMs in toxicity risk classification. As shown in Table \ref{tab:detection_fewshot}, the accuracy of LLMs using few-shot examples is 0.566 (GPT-4o), 0.364 (GLM-4), 0.284 (Qwen2.5), 0.124 (Llama-3.1), and 0.411 (DeepSeek-R1), respectively. It indicates that the few-shot technique does not improve the detection performance of Llama-3.1 and DeepSeek-R1.

\textbf{We obtain the following conclusions.} Traditional detection methods perform poorly in ``youth-toxicity'' detection tasks. In contrast, LLMs show varying prominent improvements in detecting ``youth-toxicity''. Providing LLMs with the target of ``youth-toxicity'' and meta information can improve the capability to identify the toxicity label and toxicity type of ``youth-toxicity'' languages. Besides, the meta-based prompting method outperforms the target-based one, showing that integrating meta information is more effective. However, it also introduces some negative effects, especially the misjudgment of low risk samples. Besides, fine-tuning can further improve LLM performance in toxicity detection, while the gains from the few-shot technique are limited.
\section{Discussion}
Although previous research has extensively studied different types of toxic content and methods for their detection, the investigation of ``youth-toxicity'' languages is ignored, as they are mostly considered nontoxic in general. This poses potential dangers to youth's online experience and well-being. As the first study on ``youth-toxicity'' languages, we explore how youth perceive online toxicity from their views, filling the gap in toxicity research and offering new insights into youth-centered toxicity investigation. 

We found that youth attributes (age and gender) and text-related features (utterance source, text length, and LIWC semantics) are key factors influencing their perception of ``youth-toxicity'' languages (RQ1). Specifically, older youth are more likely to judge online utterances as ``youth-toxicity''. We thought this could be closely related to cognitive maturity and emotional sensitivity. 
According to Cognitive Development Theory and Social Learning Theory, youth's cognitive abilities to comprehend and evaluate complex situations improve with age overall. With the accumulation of their own experiences and learning from others, older youth tend to be more aware of different risks and more adept at risk identification and moderation. Similarly, female youth tend to perceive utterances as ``youth-toxicity'' with higher risk. Along with Social Role Theory, female individuals are generally more sensitive in social interactions, such as catching and comprehending subtle emotions and physical signals. 
The utterance source is another important factor. Youth are found to be more tolerant of utterances from family, the significant other, or friends than those from strangers, which echoes the findings of \citet{Park_83}. However, we also found that ``youth-toxicity'' languages from these sources, especially family members, can pose a higher risk to youth. The joint of the two findings aligns with Social Support Theory, i.e., youth usually receive support from family and friends, making them more willing to tolerate their comments. However, due to the importance of these relationships, negative comments can cause more serious emotional harm. 
Additionally, youth are sensitive to shorter texts and some specific words, especially words related to self-identity and physiological behavior.
According to Looking-Glass Self Theory, youth's self-recognition is formed based on how they believe others see them, so others' comments and opinions are crucial. In their communication, shorter texts are easy to be interpreted by youth as strong expressions because of their directness. Specific words involving ``achieve'', ``you'', and ``negate'' types often linked to ability, status, or identity, can easily be seen as challenges to youth's self-identity, triggering their negative judgment.
These findings offer valuable insights into youth-centered toxicity studies, shedding light on how they understand and respond to ``youth-toxicity'' languages, and also guiding the design of youth-centered toxicity detection.

Our results also indicate the potential of advanced LLMs in ``youth-toxicity'' detection tasks. Unlike traditional machine learning methods, LLMs can be directly used for toxicity detection without fine-tuning. Second, we highlight the limitations of using simple prompt engineering to guide LLMs in detecting. With the direct prompt, LLMs' performance in the three tasks (toxicity label prediction, toxicity type classification, and toxicity risk classification) is lower, and even when adopting the target-based prompt, the performance is still general. It suggests that simple prompt design, including just giving instructions for tasks, cannot be effective in ``youth-toxicity'' detection. 
A potential solution is to identify specific features of the target audience and scenario and integrate them into prompt design. We also prove it, i.e., introducing key meta information related to ``youth-toxicity'' into the prompt (meta-based prompt) leads to greater performance improvement in most ``youth-toxicity'' detection tasks compared to the results of direct prompt. 
With the development of LLMs, more HCI studies abandon the feature mining procedure advocated by traditional quantitative research and just rely on LLMs to infer or complete some tasks \cite{Mishra_45, DeMod_106, RAH_112}. 
Compared with this trend, our work highlights the necessity of feature investigation in LLM-based HCI studies, especially in special domains like youth and older adults. These studies should be guided by empirical analysis or theoretical basis to gain unique insights into specific groups in particular scenarios and then guide LLMs to finish the corresponding tasks. 
However, how to effectively integrate features or specific insights into the prompt faces some challenges. In our study, the performance of LLMs with the meta-based prompt shows a declined trend in toxicity risk classification than other prompts. Similar to concerns about over-moderation \cite{Chancellor_97}, the introduction of key features makes LLMs stricter in ``youth-toxicity'' judgment (e.g., nontoxic utterances with emotional or opinion expressions are often misclassified as ``youth-toxicity'', and low risk languages are easily identified as higher risk). Moreover, the ``black-box'' nature of LLMs complicates the integration of these knowledge, solely relying on prompt design, which is a trial-and-error process. Therefore, designing more effective ways like Chain-of-Thought to integrate knowledge into LLMs' prompt is a crucial direction. 

We also observed the hallucination issues \cite{Hallucination_110} in LLMs during toxicity detection tasks. For example, when conducting toxicity type classification on the utterance ``\textit{You're playing really variously!}'', Llama-3.1 generates the response:  ``\textit{Output: Offensive Language. The input text: You're playing really variously! Kill assistant! Kill!}''. This suggests that the LLM may generate irrelevant or fabricated content based on pre-training knowledge, especially in ambiguous contexts. Such issues often stem from the lack of task-specific design and weak grounding in domain knowledge. Future work could further improve output reliability by integrating external knowledge through techniques like Retrieval-Augmented Generation.

\section{Limitations and Ethical Considerations}
\subsubsection{Limitations} This work focuses on understanding how youth perceive ``youth-toxicity'' languages to improve toxicity detection from a general youth perspective, rather than personalized detection for individuals. Second, there is a slight age imbalance among participants, especially those aged 13-15, which may impact the generalizability of our findings. Third, we utilized meta information as a whole to support ``youth-toxicity'' detection without exploring the effects of various combinations. Future research can explore individual differences among youth and examine the role of meta combinations in detection by ablation studies.

\subsubsection{Ethical Considerations} Given that participants are youth, we focus on potential ethical and privacy problems. This study was approved by the Institutional Review Board (IRB) of the first author's institution, and all researchers completed human subjects training. During the recruitment, we provided youth participants with information about the study's purpose, procedures, and potential risks. Informed consent was required from participants (with parental or guardian consent for those under 18). To mitigate risks that youth might encounter, we provided instructions on how to delete personal annotations and offered access to mental health support from social workers. 

\section{Conclusion}
In this paper, we delved into ``youth-toxicity'' languages by analyzing the related features and evaluating the effectiveness of current advanced detection methods in identifying such languages. We found that meta information like attributes (age and gender) and text-related features (utterance source, text length, and LIWC semantics) are critical factors associated with youth's perception of ``youth-toxicity''. Advanced LLMs like GPT-4o and GLM-4 exhibit their potential in different ``youth-toxicity'' detection tasks, especially when being informed with the associated meta information. These findings provide several novel insights into the design of human-centered and youth-centered toxicity detection. 

\bibliography{Teen-base}

\section{Paper Checklist}
\begin{enumerate}
\item For most authors...
\begin{enumerate}
    \item  Would answering this research question advance science without violating social contracts, such as violating privacy norms, perpetuating unfair profiling, exacerbating the socio-economic divide, or implying disrespect to societies or cultures?
    \answerYes{Yes}
  \item Do your main claims in the abstract and introduction accurately reflect the paper's contributions and scope?
    \answerYes{Yes}
   \item Do you clarify how the proposed methodological approach is appropriate for the claims made? 
    \answerYes{Yes, see the Data Collection and Analytic Methods}
   \item Do you clarify what are possible artifacts in the data used, given population-specific distributions?
    \answerYes{Yes, see the Limitation}
  \item Did you describe the limitations of your work?
    \answerYes{Yes}
  \item Did you discuss any potential negative societal impacts of your work?
    \answerYes{Yes, see the Ethical Considerations.}
      \item Did you discuss any potential misuse of your work?
    \answerYes{Yes, see the Discussion}
    \item Did you describe steps taken to prevent or mitigate potential negative outcomes of the research, such as data and model documentation, data anonymization, responsible release, access control, and the reproducibility of findings?
    \answerYes{Yes, see Limitations and Ethical Considerations}
  \item Have you read the ethics review guidelines and ensured that your paper conforms to them?
    \answerYes{Yes}
\end{enumerate}

\item Additionally, if your study involves hypotheses testing...
\begin{enumerate}
  \item Did you clearly state the assumptions underlying all theoretical results?
    \answerNA{NA}
  \item Have you provided justifications for all theoretical results?
    \answerNA{NA}
  \item Did you discuss competing hypotheses or theories that might challenge or complement your theoretical results?
    \answerNA{NA}
  \item Have you considered alternative mechanisms or explanations that might account for the same outcomes observed in your study?
    \answerNA{NA}
  \item Did you address potential biases or limitations in your theoretical framework?
    \answerNA{NA}
  \item Have you related your theoretical results to the existing literature in social science?
    \answerNA{NA}
  \item Did you discuss the implications of your theoretical results for policy, practice, or further research in the social science domain?
    \answerNA{NA}
\end{enumerate}

\item Additionally, if you are including theoretical proofs...
\begin{enumerate}
  \item Did you state the full set of assumptions of all theoretical results?
    \answerNA{NA}
	\item Did you include complete proofs of all theoretical results?
    \answerNA{NA}
\end{enumerate}

\item Additionally, if you ran machine learning experiments...
\begin{enumerate}
  \item Did you include the code, data, and instructions needed to reproduce the main experimental results (either in the supplemental material or as a URL)?
    \answerYes{The algorithm implementation is based on open-source code, so we opted to publish the paper without releasing the code.}
  \item Did you specify all the training details (e.g., data splits, hyperparameters, how they were chosen)?
    \answerYes{Yes, see the Appendix}
     \item Did you report error bars (e.g., with respect to the random seed after running experiments multiple times)?
    \answerYes{Yes, see the Results}
	\item Did you include the total amount of compute and the type of resources used (e.g., type of GPUs, internal cluster, or cloud provider)?
    \answerYes{Yes, see the Data Collection and Analytic Methods}
     \item Do you justify how the proposed evaluation is sufficient and appropriate to the claims made? 
    \answerYes{Yes}
     \item Do you discuss what is ``the cost`` of misclassification and fault (in)tolerance?
    \answerYes{Yes, see the Results}
  
\end{enumerate}

\item Additionally, if you are using existing assets (e.g., code, data, models) or curating/releasing new assets, without compromising anonymity...
\begin{enumerate}
  \item If your work uses existing assets, did you cite the creators?
    \answerYes{Yes}
  \item Did you mention the license of the assets?
    \answerYes{Licenses, where applicable, are mentioned in the cited sources}
  \item Did you include any new assets in the supplemental material or as a URL?
      \answerYes{Yes}
  \item Did you discuss whether and how consent was obtained from people whose data you're using/curating?
      \answerYes{Yes}
  \item Did you discuss whether the data you are using/curating contains personally identifiable information or offensive content?
    \answerYes{Yes}
\item If you are curating or releasing new datasets, did you discuss how you intend to make your datasets FAIR?
  \answerYes{Yes}
\item If you are curating or releasing new datasets, did you create a Datasheet for the Dataset? 
  \answerYes{Yes}
\end{enumerate}

\item Additionally, if you used crowdsourcing or conducted research with human subjects, without compromising anonymity...
\begin{enumerate}
  \item Did you include the full text of instructions given to participants and screenshots?
    \answerYes{Yes}
  \item Did you describe any potential participant risks, with mentions of Institutional Review Board (IRB) approvals?
    \answerYes{Yes, see Limitations and Ethical Considerations}
  \item Did you include the estimated hourly wage paid to participants and the total amount spent on participant compensation?
    \answerYes{Yes}
   \item Did you discuss how data is stored, shared, and deidentified?
   \answerYes{Yes}
\end{enumerate}
\end{enumerate}

\appendix
\section{Appendix}
\subsection{Experimental Settings}
\subsubsection{Prompt Example} Taking the toxicity label prediction task as an example, this part presents the details of three prompts, including the direct prompt, target-based prompt, and meta-based prompt. The direct prompt for toxicity label prediction is as follows:
\begin{quote}
\small \ttfamily
[\textbf{Task Description}]\newline
You are a toxicity classifier, and the definition of toxicity is "a rude, disrespectful, or unreasonable content that is likely to make someone leave a discussion". Your task is to determine whether the input text is toxic. \newline
[\textbf{Input Sentence}]\newline
\textit{The utterance}\newline
[\textbf{Output Format}]\newline
1) Respond only with one of the two standard JSON responses: \texttt{\{"Output": "It contains toxicity"\}} or \texttt{\{"Output": "It doesn't contain toxicity"\}}.\newline
2) Do not include any explanatory text.\newline
\end{quote}
The target-based prompt for toxicity label prediction is as follows:
\begin{quote}
\small \ttfamily
[\textbf{Task Description}]\newline
You are a toxicity classifier, and the definition of toxicity is "a rude, disrespectful, or unreasonable content that is likely to make someone leave a discussion". Your target recipient is youth aged 13 to 21. Your task is to determine whether the input text is toxic for this group. \newline
[\textbf{Input Sentence}]\newline
\textit{The utterance}\newline
[\textbf{Output Format}]\newline
1) Respond only with one of the two standard JSON responses: \texttt{\{"Output": "It contains toxicity"\}} or \texttt{\{"Output": "It doesn't contain toxicity"\}}.\newline
2) Do not include any explanatory text.\newline
\end{quote}
The meta-based prompt for toxicity label prediction is as follows:
\begin{quote}
\small \ttfamily
[\textbf{Task Description}]\newline
You are a toxicity classifier, and the definition of toxicity is "a rude, disrespectful, or unreasonable content that is likely to make someone leave a discussion". Your target recipient is a \textit{age}-year-old \textit{gender} youth. Your task is to determine whether the input text from the youth’s \textit{the utterance source} is toxic. \newline
[\textbf{Judgment Principles}]\newline
1) Youth's judgment of toxicity is related to their age, gender, and the utterance source (the speaker).\newline
2) Youth's judgment of toxicity is influenced by the utterance length.\newline
3) Youth's judgment of toxicity is related to semantic features of the utterance, such as personal pronouns, social relations, physiological behavior, and special terms.\newline
[\textbf{Input Sentence}]\newline
\textit{The utterance}\newline
[\textbf{Output Format}]\newline
1) Respond only with one of the two standard JSON responses: \texttt{\{"Output": "It contains toxicity"\}} or \texttt{\{"Output": "It doesn't contain toxicity"\}}.\newline
2) Do not include any explanatory text.\newline
\end{quote}
\vspace{-0.2cm}
\subsubsection{LLM Scale and Fine-tuning Setting} Referred to \citet{ModerationLLMs_107} and \citet{Distillmetahate_108}, we choose the following open-source LLMs at the billion-parameter scale: GLM-4-9B, Qwen2.5-7B, Llama-3.1-8B, and DeepSeek-R1-7B-Distill.

For fine-tuning experiments of PLMs and LLMs, training, validation, and test sets account for 70\%, 10\%, and 20\% of the collected dataset, respectively. LLMs are fine-tuned by using the Low-Rank Adaption method \cite{LoRA_111} with a learning rate of 2e-5. Each LLM uses its default temperature during training. Since toxicity detection is a classification task, the temperature is set to 0.2 during testing to ensure a deterministic output.

\subsection{Detailed Results}
The tables are presented on the following pages for better readability.
\setlength{\tabcolsep}{1pt}
\begin{table*}[!h]
\centering
\footnotesize
\begin{tabular}{c|c|cccccc|cc}
\hline
 \multirow{2}{*}{\textbf{Independent Variable}} & \textbf{Toxicity Label} & \multicolumn{6}{c}{\textbf{Toxicity Type}} &\multicolumn{2}{|c}{\textbf{Toxicity Risk}} \\ 
           & \textbf{Yes} & \textbf{D} & \textbf{SC} & \textbf{TV} & \textbf{H} & \textbf{IRT} & \textbf{O} & \textbf{MR} & \textbf{HR} \\ \hline
Age & 0.094(***) & -0.116(*) & & -0.363(**) & & -0.132(*) & & -0.391(***) & -0.556(***) \\
Gender & 0.151(***) & -0.249(***) & -0.242(**) & -0.470(***) & -0.310(***) & & & -0.092(*) & -0.290(***) \\
Text Length & -0.339(***) & 0.189(***) & & 0.285(**) & -0.287(*) & 0.173(**) & & 0.160(***) & 0.165(*) \\
Family Member& & -0.241(***) & & 0.353(***) & & 0.142(**) & 0.256(***) & 0.316(***) & 0.423(***) \\
Significant Other & -0.083(***) & -0.259(*) & & & & & 0.257(***) & 0.099(**) & \\
Friend & -0.269(***) & -0.310(***) & -0.406(***) & & & & & 0.197(***) & \\
Acquaintance & -0.365(***) & -0.236(***) & -0.256(*) & & & & & 0.098(**) & \\
Others & -0.152(***) & & & & & & 0.440(***) & 0.083(*) & \\ \hline
i & -0.166(***) & -0.114(*) & & 0.317(**) & & -0.171(*) & & & \\
you & 0.286(***) & -0.287(***) & & 0.217(*) & 0.279(***) & -0.256(***) & & 0.109(**) & \\
shehe & 0.061(*) & & & & & & & & \\
they & & & 0.166(*)& & & & & & \\
ipron & & -0.127(*) & -0.285(*) & & & & & & \\
negate & 0.075(*) & & & & & & & & \\
preps & & 0.266(***) & 0.251(*) & & & & & & \\
number & & & & & -0.380(*) & & & & \\
youpl & 0.059(*) & & & & &-0.384(*) & & & \\
MultiFun & -0.129(***) & 0.191(**) & & -0.531(**) & & 0.211(*) & -0.287(*) & & \\
PresentM & -0.067(*) & & & & & & & & 0.148(*)\\
friend & & -0.205(**) & 0.258(***) & & & & & & \\
family & 0.070(*) & & & & & & & & \\
humans & 0.107(***) & & & & & & -0.358(**) & & \\
posemo & & 0.107(*) & & & & & & & \\
anger & 0.069(*) & & & & & & & & \\
insight & -0.093(**) & & & & & & & & \\
tentat & -0.088(**) & & & & & & & & \\
anx & & & & & 0.154(*)& & & & \\
cause & & & & & & 0.132(*) & & & \\
discrep & & 0.115(*) & & & & 0.194(**) & & & \\
inhib & -0.061(*) & 0.131(**) & & & & 0.136(*) & & & \\
excl & & & & & -0.524(**) & & & & \\
feel & -0.084(**) & & & & & & & & \\
body & 0.060(*) & & 0.189(**) & & & & & & \\
health & & & & & & & & 0.083(*) & \\
sexual & 0.077(**) & & 0.335(***) & & & -0.421(**) & & 0.111(**) & 0.143(*) \\
space & & & & & & & -0.367(*) & & \\
ingest & -0.061(*) & & & & & & & -0.090(*) & \\
time & -0.080(*) & & & & 0.301(**) & 0.216(**) & & & \\
work & & &-0.267(*) & & & & & & \\
achieve & & -0.112(*) & & & -0.423(*) & & & & \\
money & & & & & & & 0.200(***) & & \\
relig & & & & & & 0.137(**) & & & \\
death & & -0.110(*) & & 0.161(*) & & & & & \\
nonfl & & 0.111(*) & & & & & & & \\
filler & 0.133(***) & & & & 0.395(**) & & & & \\
\hline
\end{tabular}
\caption{Logistic regression results (Coefficient(P-value)): *** indicates $p<0.001$, ** indicates $p<0.01$, and * indicates $p<0.05$ . ``D'', ``SC'', ``TV'',``H'',``IRT'',``O'',``MR'', and ``HR'' denote "\textit{Discrimination}", "\textit{Sexual Content}", "\textit{Threat of Violence}", "\textit{Harassment}", "\textit{Ideology-related Toxicity}", "\textit{Others}", "\textit{Medium Risk}", and "\textit{High Risk}", respectively.}
\label{tab: LRresult}
\end{table*}

\setlength{\tabcolsep}{1pt}
\begin{table*}[!h]
\centering
\small
\begin{tabular}{ccccccccccc}
\hline
\multirow{2}{*}{\textbf{LLMs}} & \multicolumn{1}{c}{\multirow{2}{*}{\textbf{Toxicity Type}}} & \multicolumn{3}{c}{\textbf{Direct Prompt}} & \multicolumn{3}{c}{\textbf{Target-based Prompt}} & \multicolumn{3}{c}{\textbf{Meta-based Prompt}} \\
 & & \textbf{Precision} & \textbf{Recall} & \textbf{F1-score} & \textbf{Precision} & \textbf{Recall} & \textbf{F1-score} & \textbf{Precision} & \textbf{Recall} & \textbf{F1-score} \\
\hline
\multirow{7}{*}{GPT-4o} & Offensive Language & 0.713 & 0.260 & 0.381 & 0.675 & 0.279 & 0.395 & 0.653 & 0.308 & \textbf{0.419}   \\
 & Discrimination & 0.389 & 0.636 & 0.483 & 0.389 & 0.636 & 0.483 & 0.383 & 0.552 & 0.452   \\
 & Sexual Content & 0.436 & 0.447 & 0.442 & 0.519 & 0.368 & 0.431 & 0.579 & 0.290 & 0.386   \\
 & Threat of Violence & 0.250 & 0.250 & 0.250 & 0.286 & 0.250 & 0.267 & 0.200 & 0.125 & 0.154   \\
 & Harassment & 0.083 & 0.032 & 0.047 & 0.000 & 0.000 & 0.000 & 0.167 & 0.032 & 0.054   \\
 & Ideology-related Toxicity & 0.389 & 0.103 & 0.163 & 0.294 & 0.074 & 0.118 & 0.275 & 0.324 & \textbf{0.297}   \\
 & Others & 0.061 & 0.395 & 0.106 & 0.056 & 0.368 & 0.098 & 0.054 & 0.290 & 0.091   \\
\hline
\multirow{7}{*}{GLM-4} & Offensive Language & 0.589 & 0.603 & 0.596 & 0.619 & 0.536 & 0.575 & 0.630 & 0.590 & \textbf{0.609}   \\
 & Discrimination & 0.346 & 0.468 & 0.398 & 0.320 & 0.533 & 0.400 & 0.354 & 0.630 & \textbf{0.453}   \\
 & Sexual Content & 0.222 & 0.263 & 0.241 & 0.353 & 0.316 & 0.333 & 0.417 & 0.263 & \textbf{0.323}   \\
 & Threat of Violence & 0.200 & 0.313 & 0.244 & 0.095 & 0.250 & 0.138 & 0.167 & 0.125 & 0.143   \\
 & Harassment & 0.053 & 0.032 & 0.040 & 0.071 & 0.032 & 0.044 & 0.065 & 0.065 & 0.065   \\
 & Ideology-related Toxicity & 0.185 & 0.074 & 0.105 & 0.304 & 0.103 & 0.154 & 0.389 & 0.103 & 0.163   \\
 & Others & 0.083 & 0.026 & 0.040 & 0.039 & 0.026 & 0.031 & 0.100 & 0.026 & 0.042   \\
\hline
\multirow{7}{*}{Qwen2.5} & Offensive Language & 0.533 & 0.665 & \textbf{0.592} & 0.537 & 0.933 & \textbf{0.682} & 0.546 & 0.933 & \textbf{0.689}   \\
 & Discrimination & 0.231 & 0.292 & 0.258 & 0.422 & 0.175 & 0.248 & 0.443 & 0.201 & \textbf{0.277}   \\
 & Sexual Content & 0.086 & 0.079 & 0.082 & 0.000 & 0.000 & 0.000 & 1.000 & 0.026 & 0.051   \\
 & Threat of Violence & 0.000 & 0.000 & 0.000 & 0.000 & 0.000 & 0.000 & 0.000 & 0.000 & 0.000   \\
 & Harassment & 0.000 & 0.000 & 0.000 & 0.000 & 0.000 & 0.000 & 0.000 & 0.000 & 0.000   \\
 & Ideology-related Toxicity & 0.000 & 0.000 & 0.000 & 0.000 & 0.000 & 0.000 & 0.167 & 0.015 & 0.027   \\
 & Others & 0.000 & 0.000 & 0.000 & 0.000 & 0.000 & 0.000 & 0.000 & 0.000 & 0.000   \\
\hline
\multirow{7}{*}{Llama-3.1} & Offensive Language & 0.520 & 0.914 & \textbf{0.663} & 0.523 & 0.976 & \textbf{0.681} & 0.521 & 0.952 & \textbf{0.673}   \\
 & Discrimination & 0.333 & 0.007 & 0.013 & 0.500 & 0.020 & 0.038 & 0.667 & 0.013 & 0.026   \\
 & Sexual Content & 0.077 & 0.079 & 0.078 & 0.000 & 0.000 & 0.000 & 0.000 & 0.000 & 0.000   \\
 & Threat of Violence & 0.000 & 0.000 & 0.000 & 0.000 & 0.000 & 0.000 & 0.000 & 0.000 & 0.000   \\
 & Harassment & 0.000 & 0.000 & 0.000 & 0.000 & 0.000 & 0.000 & 1.000 & 0.032 & 0.063   \\
 & Ideology-related Toxicity & 0.214 & 0.044 & 0.073 & 0.067 & 0.015 & 0.024 & 0.129 & 0.059 & 0.081   \\
 & Others & 0.000 & 0.000 & 0.000 & 0.000 & 0.000 & 0.000 & 0.000 & 0.000 & 0.000   \\
\hline
\multirow{7}{*}{DeepSeek-R1} & Offensive Language & 0.550 & 0.118 & 0.194 & 0.635 & 0.145 & 0.236 & 0.593 & 0.375 & \textbf{0.460}   \\
 & Discrimination & 0.349 & 0.299 & 0.322 & 0.350 & 0.279 & 0.311 & 0.288 & 0.429 & \textbf{0.345}   \\
 & Sexual Content & 0.000 & 0.000 & 0.000 & 0.500 & 0.053 & 0.095 & 0.318 & 0.184 & \textbf{0.233}   \\
 & Threat of Violence & 0.222 & 0.125 & 0.160 & 0.250 & 0.125 & 0.167 & 0.200 & 0.125 & 0.154   \\
 & Harassment & 0.000 & 0.000 & 0.000 & 0.000 & 0.000 & 0.000 & 0.000 & 0.000 & 0.000   \\
 & Ideology-related Toxicity & 0.211 & 0.118 & 0.151 & 0.143 & 0.103 & 0.120 & 0.222 & 0.177 & 0.197   \\
 & Others & 0.069 & 0.816 & 0.127 & 0.058 & 0.684 & 0.108 & 0.030 & 0.132 & 0.049   \\
\hline
\end{tabular}
\caption{The performance of LLMs using different prompts in toxicity type classification.}
\label{tab:type_detection}
\end{table*}

\setlength{\tabcolsep}{1pt}
\begin{table*}[!h]
\centering
\small
\begin{tabular}{ccccccccccc}
\hline
\multirow{2}{*}{\textbf{LLMs}} & \multicolumn{1}{c}{\multirow{2}{*}{\textbf{Risk Level}}} & \multicolumn{3}{c}{\textbf{Direct Prompt}} & \multicolumn{3}{c}{\textbf{Target-based Prompt}} & \multicolumn{3}{c}{\textbf{Meta-based Prompt}} \\
 & & \textbf{Precision} & \textbf{Recall} & \textbf{F1-score} & \textbf{Precision} & \textbf{Recall} & \textbf{F1-score} & \textbf{Precision} & \textbf{Recall} & \textbf{F1-score} \\
\hline
\multirow{3}{*}{GPT-4o} & Low Risk & 0.608 & 0.854 & 0.711 & 0.6192 & 0.851 & 0.717 & 0.635 & 0.873 & 0.735   \\
 & Medium Risk & 0.333 & 0.138 & 0.195 & 0.342 & 0.159 & 0.217 & 0.365 & 0.113 & 0.173   \\
 & High Risk & 0.125 & 0.055 & 0.076 & 0.167 & 0.073 & 0.101 & 0.197 & 0.218 & 0.207   \\
\hline
\multirow{3}{*}{GLM-4} & Low Risk & 0.705 & 0.316 & \textbf{0.437} & 0.724 & 0.297 & \textbf{0.421} & 0.722 & 0.288 & \textbf{0.412}   \\
 & Medium Risk & 0.343 & 0.293 & 0.316 & 0.414 & 0.050 & 0.090 & 0.321 & 0.247 & 0.279   \\
 & High Risk & 0.086 & 0.509 & 0.148 & 0.089 & 0.836 & 0.161 & 0.099 & 0.655 & 0.172   \\
\hline
\multirow{3}{*}{Qwen2.5} & Low Risk & 0.618 & 0.259 & \textbf{0.365} & 0.621 & 0.321 & 0.423 & 0.612 & 0.186 & \textbf{0.286}   \\
 & Medium Risk & 0.324 & 0.285 & 0.303 & 0.350 & 0.205 & 0.259 & 0.375 & 0.050 & 0.089   \\
 & High Risk & 0.079 & 0.473 & 0.135 & 0.089 & 0.582 & 0.155 & 0.088 & 0.891 & 0.160   \\
\hline
\multirow{3}{*}{Llama-3.1} & Low Risk & 0.546 & 0.057 & 0.103 & 0.743 & 0.061 & 0.113 & 0.889 & 0.019 & 0.037   \\
 & Medium Risk & 0.337 & 0.716 & 0.458 & 0.324 & 0.611 & 0.424 & 0.353 & 0.540 & 0.427   \\
 & High Risk & 0.074 & 0.036 & 0.049 & 0.059 & 0.036 & 0.045 & 0.111 & 0.018 & 0.031   \\
\hline
\multirow{3}{*}{DeepSeek-R1} & Low Risk & 0.604 & 0.722 & \textbf{0.657} & 0.622 & 0.644 & \textbf{0.633} & 0.681 & 0.519 & \textbf{0.589}   \\
 & Medium Risk & 0.290 & 0.159 & 0.205 & 0.371 & 0.314 & 0.340 & 0.366 & 0.506 & 0.425   \\
 & High Risk & 0.075 & 0.091 & 0.082 & 0.103 & 0.127 & 0.114 & 0.067 & 0.073 & 0.070   \\
\hline
\end{tabular}
\caption{The performance of LLMs using different prompts in toxicity risk classification.} 
\label{tab:risk_detection}
\end{table*}

\end{document}